\documentclass{article}

\usepackage{amsmath}
\usepackage{amssymb}
\usepackage{subcaption}
\usepackage{tabularx}
\newcolumntype{Y}{>{\centering\arraybackslash}X}
\usepackage{float}
\usepackage[hidelinks]{hyperref}
\usepackage{multirow}
\usepackage{epigraph}
\usepackage{graphicx}
\usepackage{colortbl}
\usepackage{xcolor}

\usepackage[dblblindworkshop, final]{neurips_2025}
\workshoptitle{Mechanistic Interpretability}

\usepackage[utf8]{inputenc} % allow utf-8 input
\usepackage[T1]{fontenc}    % use 8-bit T1 fonts
\usepackage{hyperref}       % hyperlinks
\usepackage{url}            % simple URL typesetting
\usepackage{booktabs}       % professional-quality tables
\usepackage{amsfonts}       % blackboard math symbols
\usepackage{nicefrac}       % compact symbols for 1/2, etc.
\usepackage{microtype}      % microtypography
\usepackage{xcolor}         % colors

\title{Some Attention is All You Need for Retrieval}
\author{
  Felix Michalak\thanks{Correspondence: For inquiries, please reach out to \texttt{felix[at]michalax[dot]de}}\\
  University of Groningen\\
  The Netherlands \\
  \texttt{k.f.michalak@student.rug.nl} \\
  \And
  Steven Abreu\\
  University of Groningen\\
  The Netherlands \\
  \texttt{s.abreu@rug.nl} \\
}

\begin{document}

\maketitle

\begin{abstract}
We demonstrate complete functional segregation in hybrid SSM-Transformer architectures: retrieval depends exclusively on self-attention layers. Across RecurrentGemma-2B/9B and Jamba-Mini-1.6, attention ablation causes catastrophic retrieval failure (0\% accuracy), while SSM layers show no compensatory mechanisms even with improved prompting. Conversely, sparsifying attention to just 15\% of heads maintains near-perfect retrieval while preserving 84\% MMLU performance, suggesting self-attention specializes primarily for retrieval tasks. We identify precise mechanistic requirements for retrieval: needle tokens must be exposed during generation and sufficient context must be available during prefill or generation. This strict functional specialization challenges assumptions about redundancy in hybrid architectures and suggests these models operate as specialized modules rather than integrated systems, with immediate implications for architecture optimization and interpretability.
\end{abstract}

%%%%%%%%%%%%%%%%%%%%%%%%%%%%%%%%%%%%%%%%%%%%%%%%%%%%%%%%%%%%
%%%%%%%%%%%%%%%%%%%%%%%%%%%%%%%%%%%%%%%%%%%%%%%%%%%%%%%%%%%%
% New structure:
% Introduction (intro/motivation/literature/contributions - max 2 pages): don't care about efficiency, want to understand hybrid models because they are used and work differently
%%% also introduce: models (refer to app for more details on models)
% Group 1: Q1 and Q2 of thesis paper (generalization of zani et al findings, jrt does not improve retrieval -> retrieval is exclusive to self-attention)
% Group 2: Q3 thesis paper (what else is dealt with in self-attention, what is missing when ablating in general tasks, general capabilities do not decline --> likely because self-attention only does retrieval)
% Group 3: Q4 thesis paper (when does the retrieval mechanism break, also gives insights into potential noise introduced by "useless" attention heads)
% (( Related work: JRT not reproduced, dejavu, etc. ))
% Conclusion: answers to diff questions + limitations
%%%%%
% TODO:
% - Steve: abstract + current section 1&2 (intro/methods) + (conclusion?)
% - Felix: core sections for diff questions/groups (incl methods)
%%%%%%%%%%%%%%%%%%%%%%%%%%%%%%%%%%%%%%%%%%%%%%%%%%%%%%%%%%%%
%%%%%%%%%%%%%%%%%%%%%%%%%%%%%%%%%%%%%%%%%%%%%%%%%%%%%%%%%%%%

\section{Introduction}\label{sec:introduction}

Hybrid SSM-Transformer architectures combine linearly scaling state-space models with quadratically scaling attention mechanisms, promising efficient processing of long sequences without sacrificing performance \citep{botev_recurrentgemma_2024,lieber_jamba_2024}. While SSMs provide computational efficiency through recurrent state evolution, they exhibit ``fuzzy memory''--capturing patterns but struggling with precise retrieval \citep{waleffe_empirical_2024}. Attention mechanisms excel at exact retrieval through direct token comparisons \citep{olsson2022context}, motivating architectures that interleave both components layer-wise. 
Despite their practical success, a fundamental question remains: \textit{do these architectural components develop overlapping capabilities for robustness, or does training induce strict functional segregation?} 
Understanding this functional organization is essential for both mechanistic interpretability and practical model design.
Overlapping capabilities would suggest redundancy, enabling architectural simplification, while strict specialization would indicate task-specific computational delegation. 

This distinction has profound implications for interpretability research. Functional segregation would enable targeted analysis of specific capabilities by studying the corresponding specialized components, facilitating a more precise mechanistic understanding. Conversely, overlapping capabilities would require analyzing complex interactions between components, complicating interpretability efforts but potentially revealing emergent computational strategies.

From a practical design perspective, understanding component specialization directly informs architecture optimization. If attention layers specialize exclusively for retrieval while SSM layers handle other language modeling functions, this knowledge enables targeted architectural modifications--such as specialized retrieval modules or attention sparsification strategies--that preserve essential capabilities while reducing computational overhead.

We present a mechanistic analysis that shines light on this question. Through systematic ablation studies across RecurrentGemma-2B/9B and Jamba-Mini-1.6, we uncover complete functional segregation: \textbf{retrieval is exclusive to self-attention layers}, while SSM components contribute nothing to this function. This finding represents a fundamental insight into how hybrid architectures organize their computational capabilities during training.

Our investigation systematically tests three falsifiable hypotheses that probe the mechanistic underpinnings of hybrid architectures:

\begin{enumerate}
\item[\textbf{H1:}] \textit{Functional Exclusivity}: Retrieval in hybrid architectures depends exclusively on self-attention layers, with SSM layers contributing no retrieval capability. Furthermore, retrieval ability in the SSM layers cannot be recovered through prompting strategies designed to improve retrieval in SSMs.

\item[\textbf{H2:}] \textit{Retrieval Specialization}: Self-attention layers specialize primarily for retrieval, such that sparsification preserving minimal retrieval-critical heads maintains general language modeling capabilities while complete ablation causes retrieval-specific failure without broad performance degradation.

\item[\textbf{H3:}] \textit{Retrieval Conditions}: Successful retrieval depends on specific mechanistic requirements: needle token exposure during the generation phase and sufficient contextual information available during either prefill or generation phases, where precise attention weight patterns constitute necessary components.
\end{enumerate}

These hypotheses directly address the central research gap by testing whether hybrid architectures exhibit functional redundancy or strict specialization. The systematic testing of these hypotheses through controlled ablation studies across multiple models provides definitive evidence for understanding hybrid model organization.

To rigorously test these hypotheses, we employ entropy-based attention sparsification inspired by \citet{liu_deja_2023} to progressively ablate attention mechanisms while monitoring retrieval performance on the Needle-In-A-Haystack (NIAH) benchmark \citep{bai2024long}. We complement this with Just Read Twice (JRT) prompting \citep{arora_just_2024} to attempt recovery of SSM-based retrieval capabilities. Additionally, we assess broader capabilities using GLUE \citep{wang2019glue} and MMLU \citep{hendrycks2020measuring} benchmarks to ensure that observed retrieval-specific effects do not indicate general performance degradation.

Our analysis reveals three critical insights:
\textbf{(1)} Across model scales and architectural variants, as few as 15\% of attention heads maintain near-perfect retrieval while complete ablation causes catastrophic failure, demonstrating that SSM layers do not contribute to retrieval.
\textbf{(2)} This sparsification preserves 84\% performance on MMLU, isolating retrieval without broadly degrading model performance, thereby revealing opportunities for more effective model design.
\textbf{(3)} Successful retrieval requires specific mechanistic conditions--needle token exposure during generation plus sufficient contextual information during prefill and generation phases--establishing precise requirements for the attention-based retrieval mechanism.

These findings provide the first definitive evidence of complete functional segregation in hybrid architectures, revealing that training induces strict specialization. This understanding enables targeted optimization strategies, more precise interpretability research, and informed design of future hybrid architectures that leverage the distinct strengths of each component type. 

\section{Methods}

\subsection{Models and Experimental Setup}

We investigate functional segregation in hybrid SSM-Transformer architectures through systematic ablation studies across three models: RecurrentGemma-2B (RG-2B), RecurrentGemma-9B (RG-9B), and Jamba-Mini-1.6 (Jamba). These models represent diverse architectural approaches to hybrid design, varying in size, attention mechanism implementation, and component organization.

RG-2B contains 2B parameters across 26 layers organized as eight repetitions of the pattern "2$\times$ SSM, 1$\times$ Attention" followed by two SSM layers, with 10 attention heads per attention layer. RG-9B extends this architecture to 9B parameters across 38 layers (12 pattern repetitions) with 16 heads per attention layer. Both RecurrentGemma models employ sliding window attention with a 2048-token window.
Jamba represents a distinct architectural approach with 51B total parameters (12B active during inference) organized across 32 layers following the pattern "3$\times$ SSM, 1$\times$ Attention, 4$\times$ SSM". Unlike RecurrentGemma, Jamba utilizes global attention with 32 heads per layer and incorporates Mixture of Experts (MoE) layers that substitute standard MLPs in every other layer. This architectural diversity enables testing whether functional segregation emerges consistently across different hybrid designs.
All experiments employ a unified experimental framework implemented through ManipuLatte\footnote{Available at \url{https://github.com/lamalunderscore/manipulatte}}, our custom library providing consistent interfaces for attention manipulation across architectures. Models were tested with batch size of one, with prompt lengths up to 4096 tokens distributed across ten lengths and ten needle depths (100 prompts total), following the protocol established by \citet{zani_contextual_2025}. 
Complete model specifications and computational resources are detailed in Appendix \ref{app:models}.

\subsection{Attention Sparsification Method}

Inspired by the contextual sparsity introduced by \citet{liu_deja_2023}, we employ a simpler entropy-based top-k sparsification to systematically ablate attention mechanisms while preserving the most informative heads. For each attention head, we calculate the entropy of attention weight distributions as:
\begin{equation}
    H(A) = -\sum_t a_t \times \log_2(a_t)
\end{equation}
where $A$ represents the attention weight vector and $a_t$ denotes the attention weight for token $t$. Lower entropy indicates more focused attention patterns, suggesting greater information content and specificity in the attention mechanism.

During sparsification, we retain only the $k$ attention heads with the lowest entropy across all layers, ablating the remaining heads by setting their corresponding values in the attention output to zero. This approach preserves heads that exhibit the most decisive attention patterns while removing those with more uniform (higher entropy) distributions. We test values of $k \in [0, N]$ where $N$ equals the total number of heads per layer (10 for RG-2B, 16 for RG-9B, 32 for Jamba), with $k=0$ representing complete attention ablation and $k=N$ representing the unmodified model.
Unless otherwise specified, we apply sparsity only during generation and keep $k=N$ fixed during prefill. See Section \ref{ss:attn-manipulation} for more details.

\subsection{Attempting to Improve SSM Retrieval with Just Read Twice}

To test whether SSM layers can recover retrieval capabilities when attention is ablated, we apply Just Read Twice (JRT) prompting \citep{arora_just_2024} for severely sparsified models ($k \in \{0, 1, 2\}$). JRT repeats the context and question to potentially activate latent retrieval mechanisms in SSM layers, providing a stringent test of functional exclusivity.

\subsection{Evaluation Benchmarks}

We employ three complementary benchmarks to isolate retrieval-specific effects from general performance degradation. The Needle-In-A-Haystack (NIAH) benchmark \citep{bai2024long} directly measures retrieval capabilities by requiring models to extract specific information (the ``needle'') from contexts filled with irrelevant text. Complete NIAH specifications and scoring rubrics are provided in Appendix \ref{app:niah}.

To assess broader language modeling capabilities, we employ GLUE \citep{wang2019glue} and MMLU \citep{hendrycks2020measuring} benchmarks using the LM-Evaluation-Harness \citep{eval-harness}. GLUE evaluates fundamental natural language understanding through nine tasks, including sentiment analysis and textual entailment (zero-shot configuration). MMLU tests advanced reasoning across 57 subjects spanning STEM, humanities, and social sciences (five-shot configuration). Both benchmarks evaluate using the log-likelihood of multiple-choice options after the prefill stage, enabling assessment of how attention sparsification affects general capabilities versus retrieval-specific functions.

\subsection{Attention Manipulation Techniques}
\label{ss:attn-manipulation}

To understand the mechanistic requirements for successful retrieval, we implement four targeted attention weight manipulation techniques applied during different inference phases. These manipulations isolate specific components of the attention mechanism to identify necessary conditions for retrieval.

\begin{figure}[h!]
    \centering
    \begin{subfigure}{0.24\linewidth}
        \centering
        \includegraphics[width=1.0\linewidth]{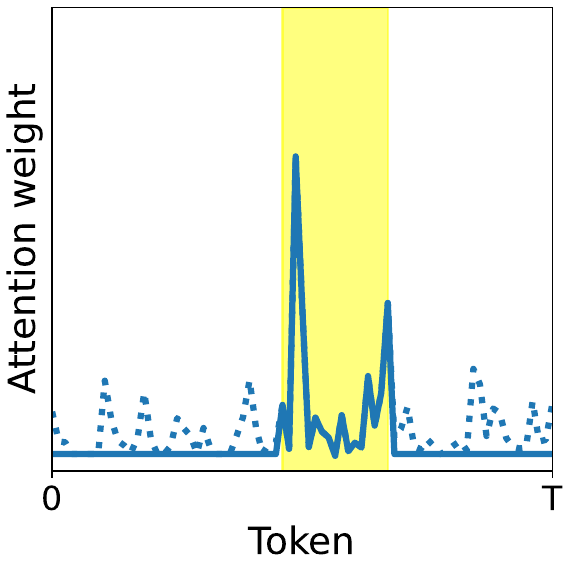}
        \caption{Only}
        \label{fig:only-expl}
    \end{subfigure}
    \begin{subfigure}{0.24\linewidth}
        \centering
        \includegraphics[width=1.0\linewidth]{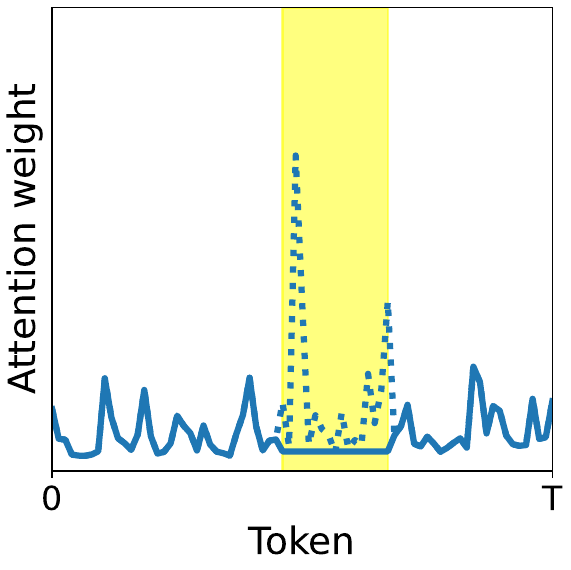}
        \caption{Omit}
        \label{fig:omit-expl}
    \end{subfigure}
    \begin{subfigure}{0.24\linewidth}
    \centering
        \includegraphics[width=1.0\linewidth]{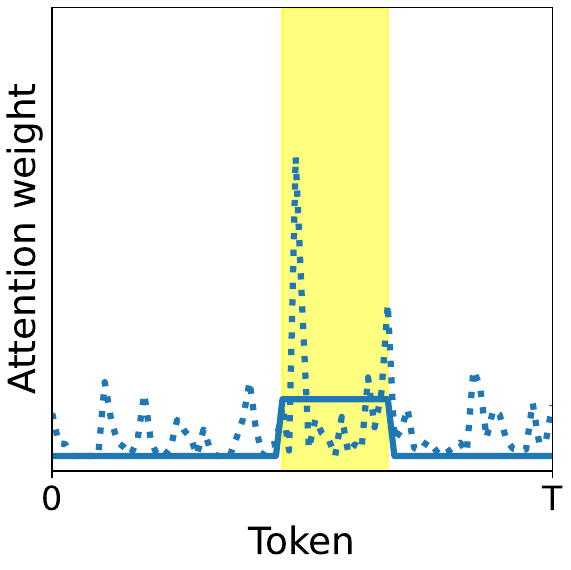}
        \caption{Binary}
        \label{fig:binary-expl}
    \end{subfigure}
    \begin{subfigure}{0.24\linewidth}
    \centering
        \includegraphics[width=1.0\linewidth]{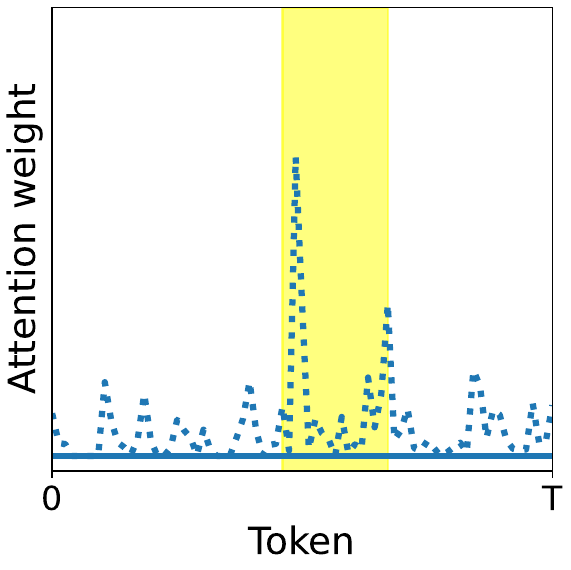}
        \caption{Null}
        \label{fig:null-expl}
    \end{subfigure}
    \caption{Simplified representation of manipulation methods. Original attention weights (dotted blue), modified weights (solid blue), and needle tokens (yellow highlight).}
    \label{fig:abl}
\end{figure}

The \textit{Only} manipulation nullifies attention weights for all tokens except needle tokens, which retain their original values, isolating the direct attention to retrieval targets. The \textit{Omit} manipulation inverts this by nullifying only needle token weights while preserving all others, testing whether contextual information alone suffices for retrieval. The \textit{Binary} manipulation extends \textit{Only} by assigning uniform average weights across needle tokens, testing whether precise weight values or merely token exposure drives retrieval. The \textit{Null} manipulation ablates all attention weights, equivalent to $k=0$ sparsification. See Figure \ref{fig:abl} for a visual representation.

% For RecurrentGemma's circular attention buffer implementation, we translate absolute token positions to circular buffer positions using:
% \[
% \text{loc}_{\text{circ}}(\text{loc}_{\text{abs}}) = \begin{cases}
%     \text{loc}_{\text{abs}} \bmod w & \text{if } \text{loc}_{\text{abs}} \ge s-w \\
%     \nexists & \text{if } \text{loc}_{\text{abs}} < s-w
% \end{cases}
% \]
% where $\text{loc}_{\text{abs}}$ represents absolute position, $\text{loc}_{\text{circ}}$ represents circular buffer position, $s$ denotes the ordinal of the generated token, and $w$ is the sliding window size (2048 tokens).

We test all combinations of prefill and generation manipulations (denoted as "Generation-Prefill", e.g., "Omit-Null" applies \textit{Omit} during generation and \textit{Null} during prefill), excluding \textit{Null} during generation alone as this equals complete sparsification. This systematic manipulation reveals the precise conditions required for the attention-based retrieval mechanism to function successfully. 

\section{Retrieval Depends Exclusively on Self-Attention}\label{q1}

Testing H1 (functional exclusivity) requires demonstrating that retrieval depends exclusively on self-attention layers with no SSM contribution, even under prompting strategies designed to enhance SSM performance. We systematically vary attention sparsification levels while monitoring retrieval performance, complemented by Just Read Twice (JRT) prompting \citep{arora_just_2024} to attempt recovery of potential latent SSM retrieval capabilities.

\subsection{Results}\label{res-q1}
% \begin{figure*}[t]
%     \centering
%     \includegraphics[width=1.0\linewidth]{img/base_NIAH_accuracy.pdf}
%     \caption{Accuracy as a function of $k$ in top-$k$ sparsification on standard NIAH for \textit{generation} and \textit{prefill versions}. Accuracy was approximated by the average score across all 100 prompts, relative to the maximum score (5). Note that scoring was neither linear nor continuous (see Table \ref{tab:scoring} for interpretation guidance). Read from right to left for increasing sparsity.}
%     \label{fig:standard_NIAH}
% \end{figure*}
\begin{figure}[h!]
    \centering
    \includegraphics[width=1.0\linewidth]{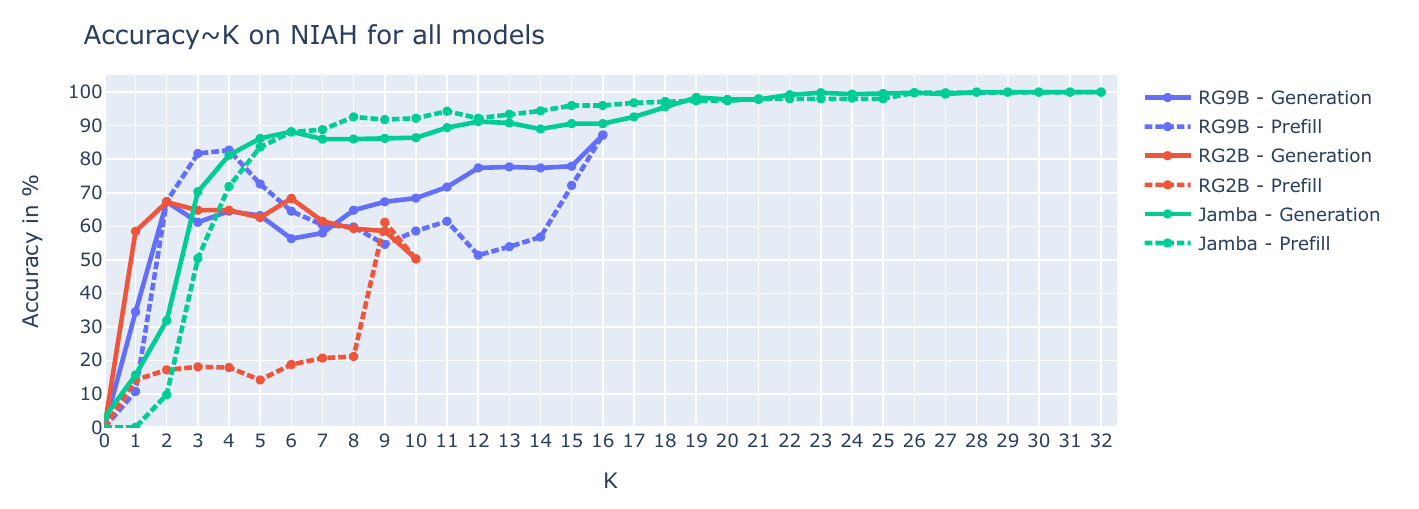}
    \caption{Accuracy as a function of $k$ in top-$k$ sparsification on standard NIAH for \textit{generation} and \textit{prefill versions}. Accuracy was approximated by the average score across all 100 prompts, relative to the maximum score (5). Note that scoring was neither linear nor continuous (see Table \ref{tab:scoring} for interpretation guidance). Read from right to left for increasing sparsity.}
    \label{fig:standard_NIAH}
\end{figure}

As seen in Figure \ref{fig:standard_NIAH}, all three models showed distinct but similar behaviors for sparsification. We refered to the model versions that are only sparsified during the generation stage as \textit{generation versions}, and to the versions that are sparsified during both stages as \textit{prefill versions}.

All models maintained high accuracy until model-specific thresholds, then showed gradual degradation before catastrophic failure. Jamba sustained 100\% accuracy until $k=27$, then declined to $k=6$ where it sharply dropped to zero. RG-9B degraded from $k=16$ to $k=3$ before failing completely at $k=2$. RG-2B's generation version remained stable until $k=1$ where it collapsed, while its prefill version immediately dropped to 20\% accuracy. Critically, all models exhibited a final tipping point ($k \in \{1,2,3,4\}$ for generation versions) where accuracy abruptly fell to zero--this threshold correlated with attention head count (32, 16, and 10 heads respectively).

Following Figure \ref{fig:jrt_all}, applying JRT did not recover retrieval capabilities at low $k$. See Appendix \ref{jrt-retrieval} for a comparison between retrieval maps for standard and JRT prompting.

\subsection{Discussion}\label{dis-q1}
Retrieval failed for all models at $k=0$, which confirmed that the findings made by \citet{zani_contextual_2025} also hold for other hybrid LLMs. However, all three models showed slightly different tipping points in generation versions, after which the accuracy started to decrease drastically: $k=4$ for Jamba, $k=2$ for RG-9B, and $k=1$ for RG-2B. Still, the overall behavior is similar across all three models. Note that our results showed the tipping point for RG-2B at $k=1$, whereas \citet{zani_contextual_2025} identified this tipping point to be around $k=2$. This difference can be attributed to our improved scoring method and prompt structure.

Regarding the prefill versions, the difference in behaviors was striking, especially prominent in the behavior of RG-2B. While the RG-2B prefill version failed even at low sparsification levels, the Jamba prefill version performed almost identically to its generation version. It seems as though the effect size of prefill sparsification reduces with increasing model size.

It is important to consider the impact of sparsification on the internals of a model. Any manipulation of the computations during a forward pass will alter the residual stream and therefore yield, compared to the original activations, an imperfect version of all downstream activations. This means that every layer activation after the first manipulation is imperfect, and every further manipulation likely increases the magnitude of the effect. This also means that every query-, key- and value-projection is imperfect. Since the RG models used a cache for the values of key- and value-projections (kv-cache) to optimize inference time, the query-projection will use the faulty cached key- and value-projections for every subsequent forward pass during that inference run, increasing the impact of the manipulation. Especially important for this cache is the prefill stage, as the cache gets filled with all projections of the prompt tokens. Jamba was not set up to use this caching mechanism. Hence, we would expect the RG models to be impacted more by prefill sparsification.

Since all models failed to retrieve the needle at $k=0$, and applying JRT did not recover this retrieval capability, this provides further strong evidence that retrieval in hybrid LLMs is exclusively implemented through self-attention layers. This also further solidifies the hypothesis that during training, only self-attention layers learn the retrieval function, so much so that SSM layers do not even develop the common \textit{fuzzy memory} \citep{waleffe_empirical_2024}.

% To investigate these training dynamics, one could identically train a pure SSM and a hybrid version of the same SSM that only adds self-attention layers. After it was confirmed that the hybrid model indeed failed retrieval tasks with ablated self-attention, the state transition matrices of the SSM layers of both models should be compared. Since the pure SSM layers will have learned a retrieval function (at least for fuzzy memory), but the layers of the hybrid SSM will not, contrasting the state transition matrices could uncover the retrieval mechanism in the pure SSM layers. Uncovering the retrieval mechanism used in SSMs would provide an example approach for mechanistic SSM analysis, furthering the field of Mechanistic Interpretability.
Future work could compare state transition matrices between pure and hybrid SSM variants to uncover how retrieval mechanisms differ when attention is available versus absent, potentially revealing fundamental SSM retrieval strategies.

\section{Self-Attention Specializes for Retrieval}\label{q3}

Testing H2 (retrieval specialization) requires demonstrating that self-attention layers specialize primarily for retrieval, such that sparsification preserving minimal retrieval-critical heads maintains general capabilities. We evaluate this through systematic benchmarking on GLUE and MMLU at different sparsification levels.

We test three configurations on Jamba: base (unmodified), optimally sparsified ($k=5$, the minimal prefill sparsification maintaining near-perfect retrieval from Section \ref{q1}), and fully ablated ($k=0$, complete attention removal). This comparison isolates the impact of attention sparsification on general versus retrieval-specific capabilities.
We apply sparsity during prefill, as the benchmark answers are evaluated by log-likelihood without generation.

\subsection{Results}

\begin{figure}[h!]
    \centering
    \includegraphics[width=0.49\linewidth]{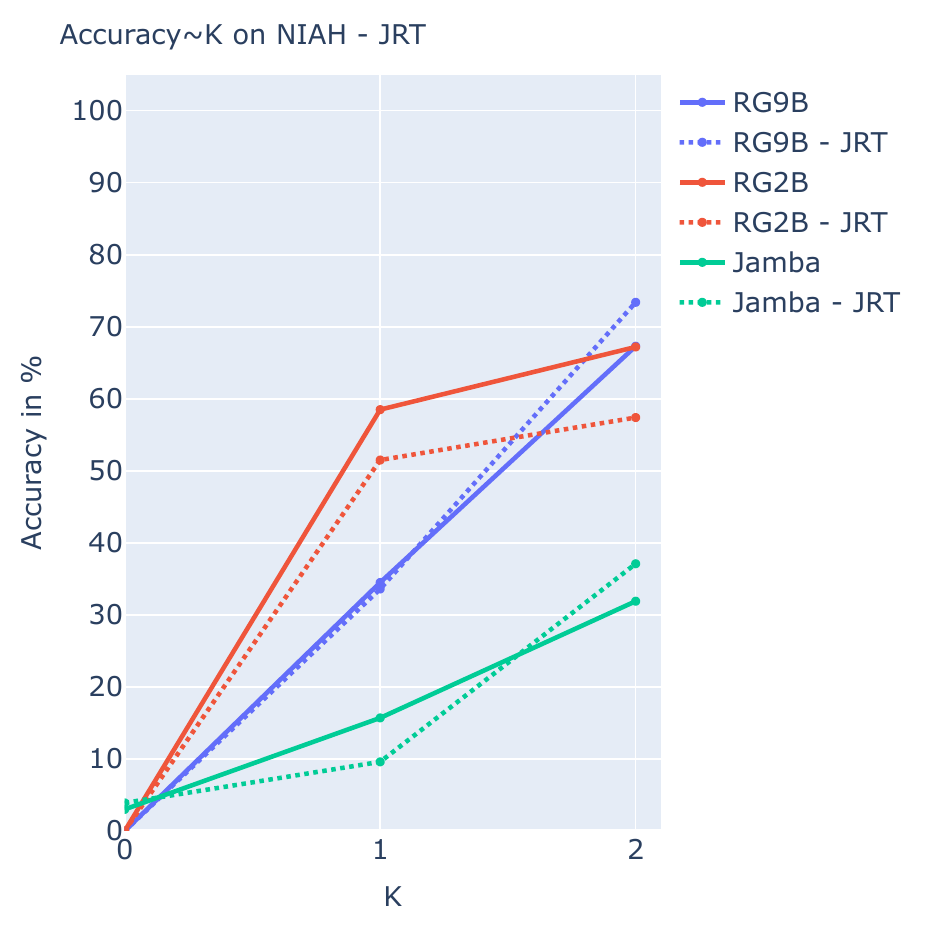}
    \includegraphics[width=0.49\linewidth]{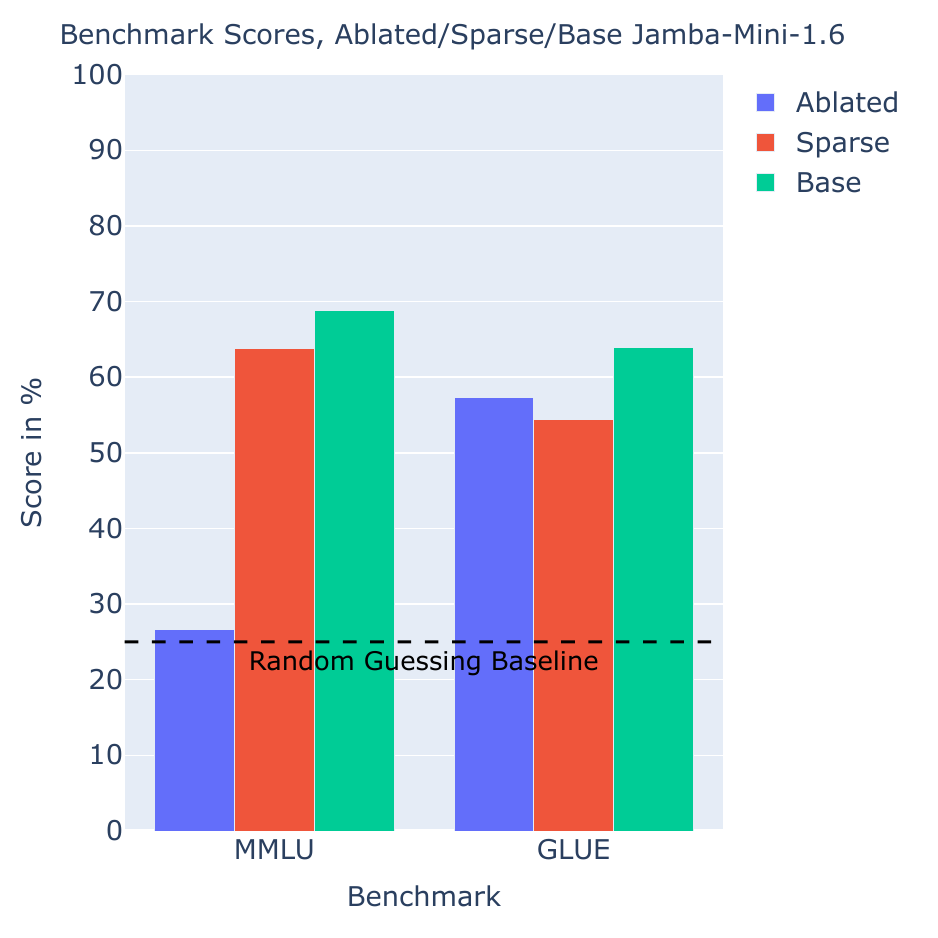}
    \caption{\textbf{Left}: Accuracy as a function of $k$ in top-$k$ sparsification on NIAH with and without JRT applied, for all models in the generation version. Accuracy was approximated by the average score across all 100 prompts, relative to the maximum score (5). Note that scoring was neither linear nor continuous (see Table \ref{tab:scoring} for interpretation guidance). Read from right to left for increasing sparsity. 
    \textbf{Right:} Scores for Jamba-Mini-1.6, in an ablated configuration ($k=0$), a sparse configuration ($k=5$), and the base configuration, on GLUE and MMLU.}
    \label{fig:jrt_all}
    \label{fig:total_benchmark}
\end{figure}

% Figure \ref{fig:total_benchmark} shows the average achieved scores per benchmark, per model configuration. See Appendix \ref{lme} for subtask results.
Figure \ref{fig:total_benchmark} shows the average achieved scores per benchmark, per model configuration. See Appendix \ref{lme} for subtask results.

% On GLUE, the base configuration of Jamba performed best, but the ablated and sparsified versions only scored slightly lower (6.6 and 9.1 points, respectively). Although only marginally, the ablated version scored higher than the sparse version (2.8 points).
On GLUE, the base configuration of Jamba performed best, but the ablated and sparsified versions only scored slightly lower (6.6 and 9.1 points, respectively). Although only marginally, the ablated version scored higher than the sparse version (2.8 points).
% On MMLU, the base configuration performed best again. The sparse version only scored six points below the base configuration, whereas the ablated version scored around 42 points below the base configuration.
On MMLU, the base configuration performed best again. The sparse version only scored six points below the base configuration, whereas the ablated version scored around 42 points below the base configuration.

% For the more challenging MMLU benchmark, the sparse performance only declined 7.4\% (relatively), despite the same configuration showing a more substantial 16.4\% decrease in NIAH (see Figure \ref{fig:standard_NIAH}). On the one hand, since the average decrease on MMLU is much lower than the decrease in retrieval capabilities, this suggests that the self-attention layers do not host other critical and exclusive capabilities beyond retrieval. On the other hand, the ablation ($k=0$) completely failed on MMLU, dropping close to the random guessing baseline, which could be an indicator that self-attention layers play a role outside of retrieval, and that their contribution is crucial to the model's inner workings. However, the GLUE results contradicted this theory by showing that the ablated configuration and the sparse configuration score similarly (and close to the base configuration) on GLUE, with the ablated configuration even coming out on top on average.

% \subsection{Discussion}
\subsection{Discussion}
% The results provide evidence that the retrieval capability of self-attention layers can be isolated through sparsification. However, the effect of sparsification depends significantly on the complexity and nature of the evaluated task.
The results provide evidence that the retrieval capability of self-attention layers can be isolated through sparsification. However, the effect of sparsification depends significantly on the complexity and nature of the evaluated task.

% The big difference in the effect of sparsification and ablation between MMLU and GLUE highlights the necessity for retrieval capabilities for more difficult language modeling tasks, whereas general language capabilities do not depend on retrieval to such a degree.
The big difference in the effect of sparsification and ablation between MMLU and GLUE highlights the necessity for retrieval capabilities for more difficult language modeling tasks, whereas general language capabilities do not depend on retrieval to such a degree.

% Still, the finding that the sparsified configuration performed worse than the ablated configuration hints at a suboptimal sparsification method. There are two likely explanations for this observation. First, the sparsification metric may be inadequately selecting which attention heads to preserve and ablate, preserving heads that potentially introduce noise while ablating the ones that are crucial for retrieval. Second, building on the first explanation, an incomplete retrieval mechanism may be worse than no retrieval at all, as a broken retrieval function that was learned to be crucial for certain tasks could lead to inconsistent or misleading information flow. The effect of such a broken retrieval function could range from simply introducing random noise (as mentioned in the first explanation) to systematically moving the activation vector in a wrong direction in the latent space, potentially pushing values out of distribution.
Still, the sparsified configuration performing worse than the ablated configuration in GLUE hints at a suboptimal sparsification method. Ablating the wrong attention heads can lead to broken retrieval functions, effectively introducing noise or even facilitating a misleading information flow.
% The findings from GLUE highlight the necessity for a more sophisticated sparsification method. Instead of investigating attention heads behaviorally, \citet{olsson2022context} structurally investigated attention heads through an offline analysis that examines the weights of different circuits in the self-attention architecture. Such an analysis could provide insights into which heads are implementing which kind of capabilities and functions before running inference. This way, instead of dynamically choosing the attention heads to ablate at run time, a model could be sparsified statically, based on the characteristics of different heads, and with more care.
Instead of investigating attention heads behaviorally, future iterations could structurally investigate attention heads through an offline analysis that examines the weights of different circuits in the self-attention architecture, as performed by \citet{olsson2022context}. Such an analysis could provide insights into which heads are implementing which kind of capabilities and functions before running inference. This way, instead of dynamically choosing the attention heads to ablate at run time, a model could be sparsified statically, based on the characteristics of different heads, and with more care.

\section{Mechanistic Requirements for Retrieval}\label{q4}
Testing H3 (retrieval conditions) requires identifying the specific mechanistic requirements for successful retrieval. We systematically manipulate attention weights to isolate necessary conditions: needle token exposure and contextual information availability during different inference phases.

We apply the four manipulation techniques described in Section \ref{ss:attn-manipulation} (\textit{Only}, \textit{Omit}, \textit{Binary}, \textit{Null}) to RG-2B without sparsification ($k=N$), testing all combinations of prefill and generation manipulations. This approach reveals which attention components are mechanistically necessary for retrieval.

\subsection{Results}

\begin{table}[h!]
\hspace{-0.5cm}
    \setlength{\extrarowheight}{3pt}
    \begin{tabularx}{1.02\linewidth}{cY|Y|Y|Y|Y}
        \multicolumn{2}{c}{} & \multicolumn{4}{c}{Generation} \\
        \multicolumn{2}{c|}{} &
                   Keep &  Omit&   Only &   Binary \\
        \cline{2-6}
        \multirow{5}{*}{\rotatebox{90}{Prefill}} &
        Keep &     50.3\% & 3.7\% & \textbf{\underline{85.7\%}} & 3.7\%\\
        \cline{2-6}
        &Omit&     63.0\% & 0.0\% & 53.0\% & 0.0\%\\
        \cline{2-6}
        &Only&     \textbf{70.6\%} & 0.1\% & 18.2\% & 0.1\%\\
        \cline{2-6}
        & Binary & 0.0\%  & 0.0\% & 0.0\%  & 0.0\%\\
        \cline{2-6}
        & Null &   \underline{68.8\%} & 0.0\% & 17.0\% & 0.0\%\\
    \end{tabularx}
    \caption{NIAH results for all tested manipulation combinations on RecurrentGemma-2B.Marked as \textbf{\underline{overall best}}, \textbf{second best} and \underline{third best}}.
    \label{tab:man_res}
\end{table}

Table \ref{tab:man_res} gives an overview of the accuracies scored in NIAH for different combinations of prefill and generation manipulation. Omitting the needle tokens during generation led to a drastic drop in accuracy, with Omit-Keep reaching 3.7\% accuracy, and all other combinations scoring zero or close to zero percent accuracy. Applying the Binary method during generation yielded the same accuracies as applying the Omit method; all combinations applying Binary during prefill scored zero percent accuracy. The combinations Keep-Omit, Keep-Only, and Keep-Null all improved the accuracy relative to the base configuration, with 63.0\%, 70.6\%, and 68.8\%, respectively. When only keeping the needle tokens during prefill, not manipulating the prefill, or omitting the needle tokens also improved the scores relative to the base configuration, with 85.7\% and 53.0\%, respectively. Only-Keep reached the highest score across all combinations. Note that every sparsification combination that did not severely reduce retrieval capabilities yielded better scores than the base model (Keep-Keep).

\begin{figure}
    \centering
    \begin{subfigure}{0.29151\linewidth}
        \centering
        \includegraphics[width=1.0\linewidth]{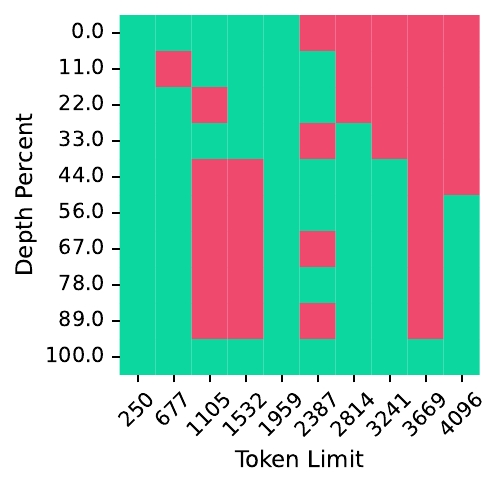}
        \caption{Keep-Omit}
        \label{fig:keep-omit}
    \end{subfigure}
    \begin{subfigure}{0.273837\linewidth}
        \centering
        \includegraphics[width=1.0\linewidth]{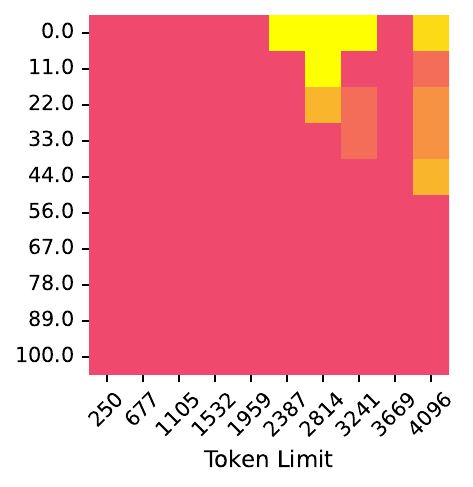}
        \caption{Omit-Keep}
        \label{fig:omit-keep}
    \end{subfigure}
    \begin{subfigure}{0.33401\linewidth}
        \centering
        \includegraphics[width=1.0\linewidth]{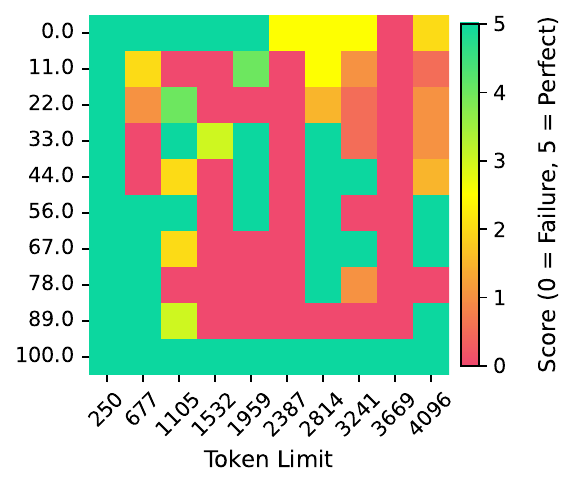}
        \caption{Keep-Keep (Base)}
        \label{fig:keep-keep}
    \end{subfigure}
    \caption{Retrieval maps for RG-2B on NIAH with the manipulations Keep-Omit, Omit-Keep, and Keep-Keep applied. The x-axis shows the used prompt length, the y-axis the depth of the needle, 0\% being the very end of the prompt and 100\% being the very beginning of the prompt.}
    \label{fig:manipulation}
\end{figure}
For a complete overview of all retrieval maps, see Appendix \ref{attention-manipulation}.

\subsection{Discussion}
The results suggest that successful retrieval necessitates, firstly, exposure of the needle tokens during generation, and secondly, exposure of tokens that provide sufficient context either during prefill or during generation. The first point is obvious from the results when applying Omit during generation. Every combination, including Omit-Keep, fails retrieval. To deduce the second point, the key combinations to look at are Only-Omit and Only-Only, as well as Keep-Omit, Keep-Only, and Keep-Null. Only-Omit shows that, even when only exposing the needle tokens during generation, only exposing the structure around the needle during prefill suffices to facilitate successful (above baseline) retrieval capabilities; when not exposing the structure during prefill, like in Only-Only, retrieval fails. The other three combinations, especially Keep-Null, show that the tokens exposed during prefill are irrelevant, likely because enough tokens are exposed during generation.

Additionally, the low accuracy scores when applying Binary showed that the values of the attention weights are important for successful retrieval. It is not enough to expose necessary tokens more than unnecessary ones, but the exact structure of and relations between the attention weights matter.

Investigating the retrieval maps of different combinations revealed further details, refining the previous hypothesis. The retrieval maps for Keep-Omit (Figure \ref{fig:keep-omit}) and Omit-Keep (Figure \ref{fig:omit-keep}) complement each other to make up an improved version of Keep-Keep (Figure \ref{fig:keep-keep}). This means that Omit-Keep, which was previously labeled as a retrieval failure, was not a failure but only a partial success.

% The phenomenon we can see here is likely related to the 2048 token sliding window attention that the RG models employ. Observe that the triangle of partially successful retrievals in the top right corner of Figure \ref{fig:omit-keep} starts at a prompt length of 2387, which is the first tested prompt length after 2048 tokens. For all unsuccessful combinations of prompt length and needle depth, the needle was not included in the sliding window.
% This is only possible because the RG models only employ sliding window attention during the autoregressive sampling, but not during prefill. The fact that there is even partially successful retrieval means that either the first output token (which is generated by prefill) carries a large enough bias, or the kv-cache generated during prefill encoded enough implicit information to make the model partially retrieve correct information despite this information not being exposed to the self-attention mechanism during further inference steps. It also means that the retrieval function learned by the model generalized beyond 2048 tokens. The retrieval maps in (show that retrieval capabilities beyond the sliding window in RG2B are strictly dependent on all necessary tokens (needle and sufficient context) being exposed during the prefill stage, as neither Omit-Omit nor Omit-Only show this behavior. See Figure \ref{fig:manip_total} for the retrieval maps of all combinations shown in Table \ref{tab:man_res}.
The partial retrieval success in Figure \ref{fig:omit-keep} aligns with RecurrentGemma's 2048-token sliding window--successful retrievals occur only when needles fall within this window during generation. Retrieval beyond the window requires prefill exposure, suggesting the kv-cache preserves implicit retrieval information even when tokens exit the active attention window. See Figure \ref{fig:manip_total} for complete retrieval maps.

Since the Binary method yielded such detrimental results, we examined the outputs of corresponding combinations. The outputs showed that when Binary was applied during prefill, the generated tokens were mostly nonsensical. When Binary was applied during generation, the generated tokens made sense, but were off topic. See Appendix \ref{memos} for a selection of pearls of wisdom generated by confused hybrid models.

Applying Binary during prefill likely has a similar effect on the kv-cache as sparsifying during the prefill stage (see Section \ref{dis-q1}), producing nonsensical tokens during generation. This would also explain why applying Binary only during generation did not result in such catastrophic failures, as the kv-cache was prefilled with unmanipulated keys and values.

Overall, the Binary method was not thoroughly investigated before employing it, but rather it was implemented in a way that most efficiently applied the idea of unifying attention weights across a range of tokens. For example, upon examining attention weights for the first activations in the generation stage, it became obvious that the range of the needle tokens always started with a large peak in the weights on the exact token that would have to be predicted (also visible in Figure \ref{fig:abl}). The behavior of the attention weights throughout the generation process was not further investigated. This peak likely moves across the range of needle tokens, always being located on the token corresponding to the correct prediction. A behavior like this would mean that the method used to apply the idea of unifying and binarizing attention weights was inadequate. This limitation could be overcome in future research.

\section{Conclusion}
% This paper investigates the role of self-attention mechanisms in hybrid language models through systematic experiments on RecurrentGemma-2B, RecurrentGemma-9B, and Jamba-Mini-1.6, addressing four key research questions about retrieval capabilities and sparsification potential.
% Our experiments demonstrate that retrieval in hybrid LLMs depends exclusively on self-attention layers. Ablation of self-attention resulted in complete retrieval failure across all tested models, confirming that SSM layers do not contribute to retrieval, as even the \textit{fuzzy memory} observed in pure SSMs \citep{waleffe_empirical_2024} is absent. The Just Read Twice method \citep{arora_just_2024} failed to recover any retrieval capabilities in sparsified models, further confirming this exclusive dependence.
% The retrieval capabilities of self-attention could be isolated without detrimental effects on the model, as shown by our benchmarks on different model versions. Still, the effect of self-attention sparsification was dependent on the tested task.
% We showed that successful retrieval requires two conditions: to-be-retrieved tokens must be exposed during generation, and sufficient context must be available during prefill or generation. Furthermore, the exact attention weight values matter, which means that binarizing attention weights post-hoc leads to retrieval failure.
We established complete functional segregation in hybrid SSM-Transformer architectures: retrieval depends exclusively on self-attention layers while SSMs handle general language capabilities. Across RecurrentGemma and Jamba models, attention ablation caused catastrophic retrieval failure (0\% accuracy) with no SSM compensation, even under specialized prompting. Yet sparsifying to just 15\% of attention heads preserved near-perfect retrieval and 84\% MMLU performance, revealing precise functional specialization.
Our entropy-based sparsification method showed suboptimal performance in some cases, indicating the need for more sophisticated approaches such as structural circuit analysis of the self-attention mechanism, as was performed by \citet{olsson2022context}.

% These findings suggest that hybrid architectures can be optimized through targeted attention sparsification without significant performance degradation on non-retrieval tasks. The exclusive specialization of attention for retrieval indicates that these components could be designed specifically for this function, potentially reducing computational overhead and increasing explainability.

This functional segregation in hybrid models enables immediate practical optimizations--targeted sparsification, specialized retrieval modules, and component-specific deployment strategies. Future work may explore whether this specialization emerges from architectural constraints or training dynamics, and whether lightweight retrieval-specific mechanisms could replace full attention layers. Understanding this functional division advances both mechanistic interpretability and efficient architecture design, suggesting hybrid models operate as task-specific specialists rather than fully integrated systems.

% This work establishes that self-attention serves as a specialized retrieval module in hybrid models, while SSM layers handle general language capabilities. Understanding this division of functionalities is crucial for designing efficient hybrid architectures that achieve transformer-level performance with improved computational efficiency, for developing more explainable substitutes to self-attention, and to understand the magic behind transformers. The exclusive dependence on attention for retrieval provides clear guidance for optimizing these architectures through targeted sparsification strategies and points out pathways for future research.

%%%%%%%%%%%%%%%%%%%%%%%%%%%%%%%%%%%%%%%%%%%%%%%%%%%%%%%%%%%%

\begin{ack}
This research was conducted as part of Felix Michalak's Bachelor's thesis at the University of Groningen under the supervision of Steven Abreu. Felix Michalak was responsible for the entirety of the code, experiments, result analysis, and discussion. Steven Abreu helped shape and streamline the project, as well as assisted with the writing process.

\end{ack}

%%%%%%%%%%%%%%%%%%%%%%%%%%%%%%%%%%%%%%%%%%%%%%%%%%%%%%%%%%%%

\bibliographystyle{bibstyle}
\bibliography{neurips_references}

\clearpage

\appendix
\section{Model Specifications}\label{app:models}

We used three different models: RecurrentGemma-2B (RG-2B), RecurrentGemma-9B (RG-9B), and Jamba-Mini-1.6 (Jamba).

RG-2B was used by \citet{zani_contextual_2025} as a comparatively small hybrid LLM that is easy to run. We used this model to provide comparability to previous results. This model consists of 2.03B parameters (excluding embedding parameters) spread across 26 layers, being eight repetitions of the pattern "2$\times$ SSM, 1$\times$ Attention" followed by two SSM layers. Each attention layer hosts ten attention heads.

RG-9B was used as a model of a larger size that is comparable to RG-2B. RG-9B and RG-2B are of the same structure and architecture, just that RG-9B is bigger in scale with 7.53B parameters spread across 38 layers, being 12 repetitions of the pattern "$2\times$ SSM, $1\times$ Attention" followed by two SSM layers. The attention layers of RG-9B host 16 heads each.

Both RG models use sliding window attention with a sliding window size of 2048.

Jamba was used as a model of different architecture to compare to the RG models. Jamba uses global attention and Mixture of Experts (MoE) layers that systematically substitute Multilayer Perceptrons (MLPs). MoE layers are sparse alternatives to MLPs that only use a fraction of parameters during inference, effectively reducing the memory footprint of a model. Jamba consists of 51.03B total parameters (excluding embedding parameters), of which only 12B are active during inference, spread across 32 layers in a pattern of "3$\times$ SSM, 1$\times$ Attention 4$\times$ SSM". Each attention layer hosts 32 attention heads.

These three models built a diverse basis for experimental results, differing in their parameter sizes, attention implementation, and surrounding architecture.

All experiments were conducted on a high-performance cluster, utilizing a maximum of two AMD Zen 3 EPYC 7763 processors, eight NVIDIA A100 (40 GB HBM2) graphics cards, and 1024 GB of memory.

\section{NIAH Benchmark Implementation Details}\label{app:niah}

The Needle-In-A-Haystack (NIAH) benchmark is built on the implementation by \citet{bai2024long}. The task in NIAH is to retrieve a specific sentence from a context that is filled with irrelevant text. The to-be-retrieved information is called the \textit{needle}, and the whole context, including the irrelevant text and the needle, is called the \textit{haystack}. The benchmark evaluates the ability of a model to retrieve the needle from the haystack for multiple prompts of different lengths. Through strategic prompt generation, this benchmark can yield retrieval maps that unveil the retrieval performance of a model, depending on the position of the needle and the size of the haystack. See Figure \ref{fig:manipulation} for examples of such a retrieval map. Following \citet{bai2024long}, we used the same collection of essays from Paul Graham\footnote{All essays can be found in their original form at \href{https://www.paulgraham.com/articles.html}{https://www.paulgraham.com/articles.html}, and are also part of the GitHub repository for this project at \href{https://github.com/lamalunderscore/retrieval-in-hybrid-ssms}{https://github.com/lamalunderscore/retrieval-in-hybrid-ssms}.}
to generate the haystack, as well as the same needle.

The needle was "The best thing to do in San Francisco is eat a sandwich and sit in Dolores Park on a sunny day". The retrieval question was "What is the best thing to do in San Francisco?".

We chose a maximum prompt length of 4096 tokens. For each tokenizer, 100 prompts were generated: ten prompt lengths at ten depths.

As an improvement over the scoring method used by \citet{zani_contextual_2025}, which was a simple binary string matching evaluation, we implemented granular scoring. While still based on string matching, partial matches were allowed to yield partial scores. For each keyword of the needle that was present in the output string, the score was increased. The scores for all partial keywords added up to three. If the full needle string matched, the score was set to five. While examining outputs from test experiments on Jamba, we stumbled across a curious phenomenon: the grammatically imperfect needle was predicted with corrected grammar, thereby only scoring three points. To visualize this phenomenon, we added a rule: if the needle was predicted with corrected grammar, the score was set to four. See Table \ref{tab:scoring} for concrete strings and their scoring behavior.

 \begin{table}[h!]
     \setlength{\extrarowheight}{3pt}
     \begin{tabularx}{1.0\linewidth}{X|c}
        \textbf{Keyword} & \textbf{Score}  \\ \hline
        "eat a sandwich"& increase by 1.0 \\ \hline
        "Dolores Park" & increase by 0.5 \\ \hline
        "sit in Dolores Park" & increase by 0.5 \\ \hline
        "sunny day" & increase by 1.0 \\ \hline
        "is to eat a sandwich and sit in Dolores Park on a sunny day" & set to 4.0 \\ \hline
        "The best thing to do in San Francisco is eat a sandwich and sit in Dolores Park on a sunny day" & set to 5.0\\
     \end{tabularx}
     \caption{Granular scoring system for NIAH evaluation. Keyword strings used in the evaluation of retrieval success, and their influence on the score. Notice the grammatical imperfection in the full needle string (last row).}
     \label{tab:scoring}
 \end{table}

\subsection{Prompt templates used in NIAH}\label{templates}
For base models that were not instruction-tuned, we used the following template to style the prompts:

\begin{verbatim}
CONTEXT:
<haystack>

QUESTION:
<retrieval question>

ANSWER: Here is the most relevant sentence in the context:
\end{verbatim}

For instruction-tuned models, we changed the template to:

\begin{verbatim}
CONTEXT:
<haystack>

QUESTION:
<retrieval question> Output the most relevant sentence in the context, word by word!
\end{verbatim}

\section{Comparing retrieval maps for JRT and standard NIAH}\label{jrt-retrieval}
\begin{figure}[h!]
\centering
\begin{subfigure}{0.3\linewidth}
    \centering
    \includegraphics[width=0.8\linewidth]{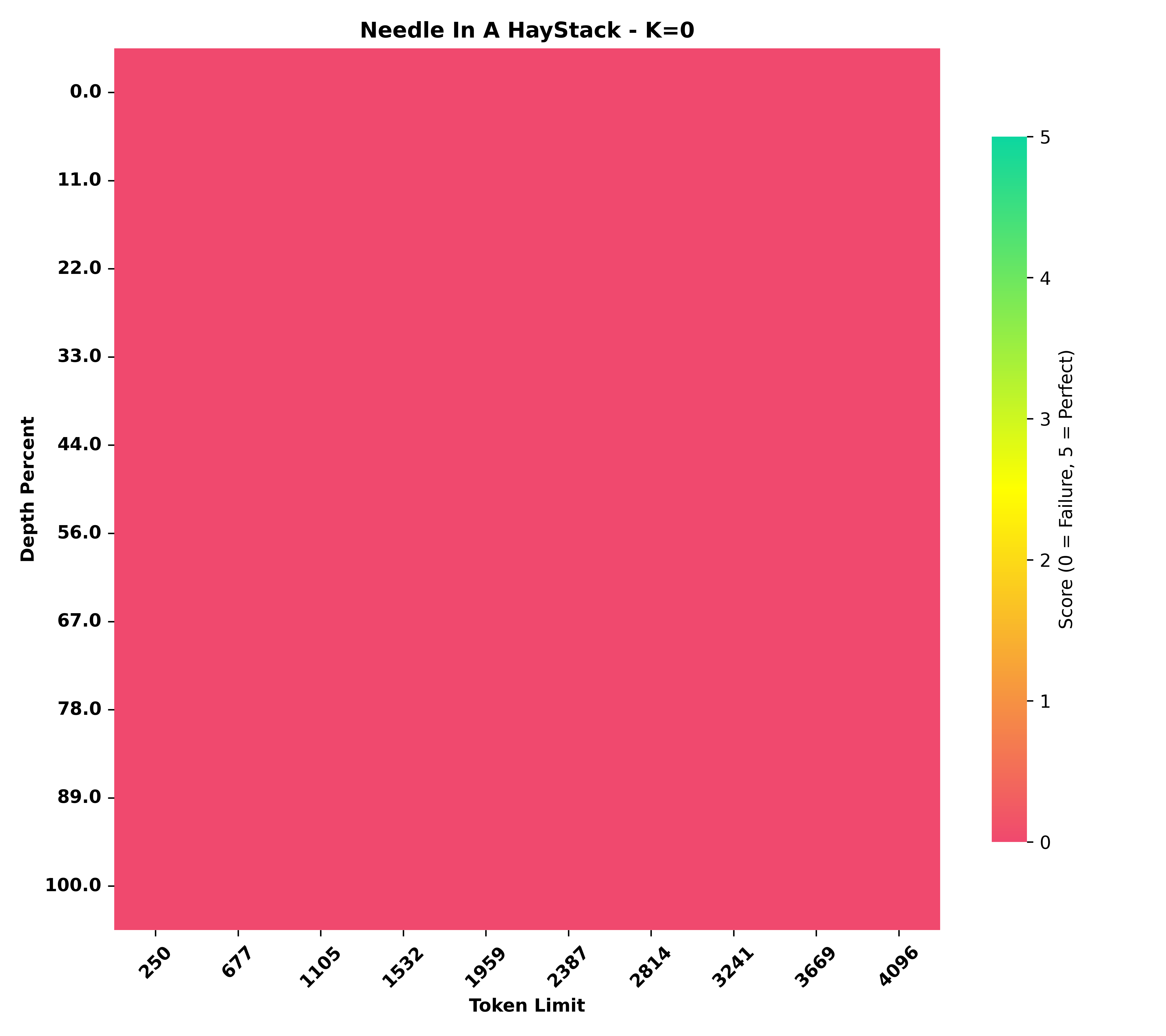}
    \caption{RG-2B - Base}
    \label{fig:rg2b_k0}
\end{subfigure}
\begin{subfigure}{0.3\linewidth}
    \centering
    \includegraphics[width=0.8\linewidth]{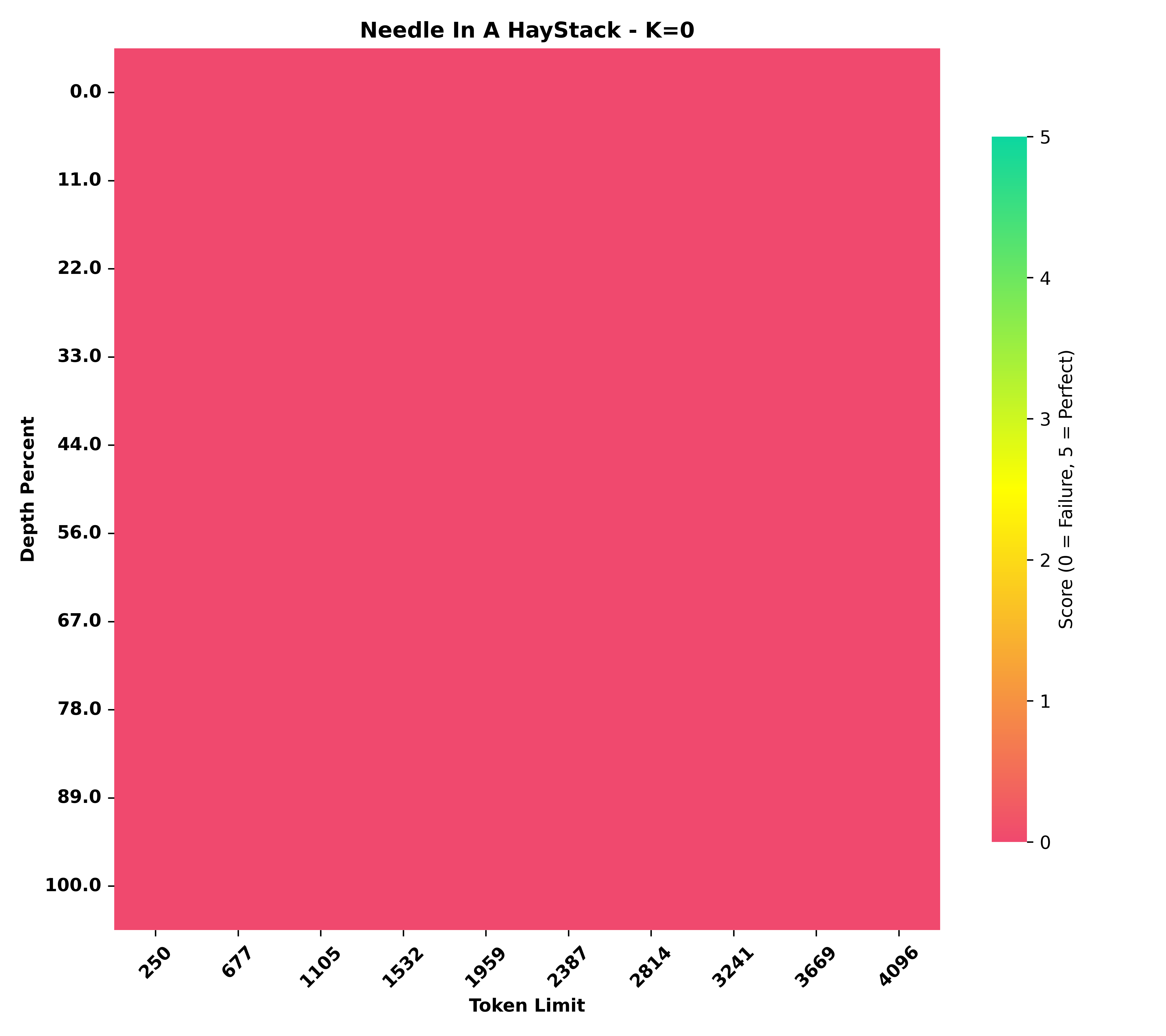}
    \caption{RG-9B - Base}
    \label{fig:rg9b_k0}
\end{subfigure}
\begin{subfigure}{0.3\linewidth}
    \centering
    \includegraphics[width=0.8\linewidth]{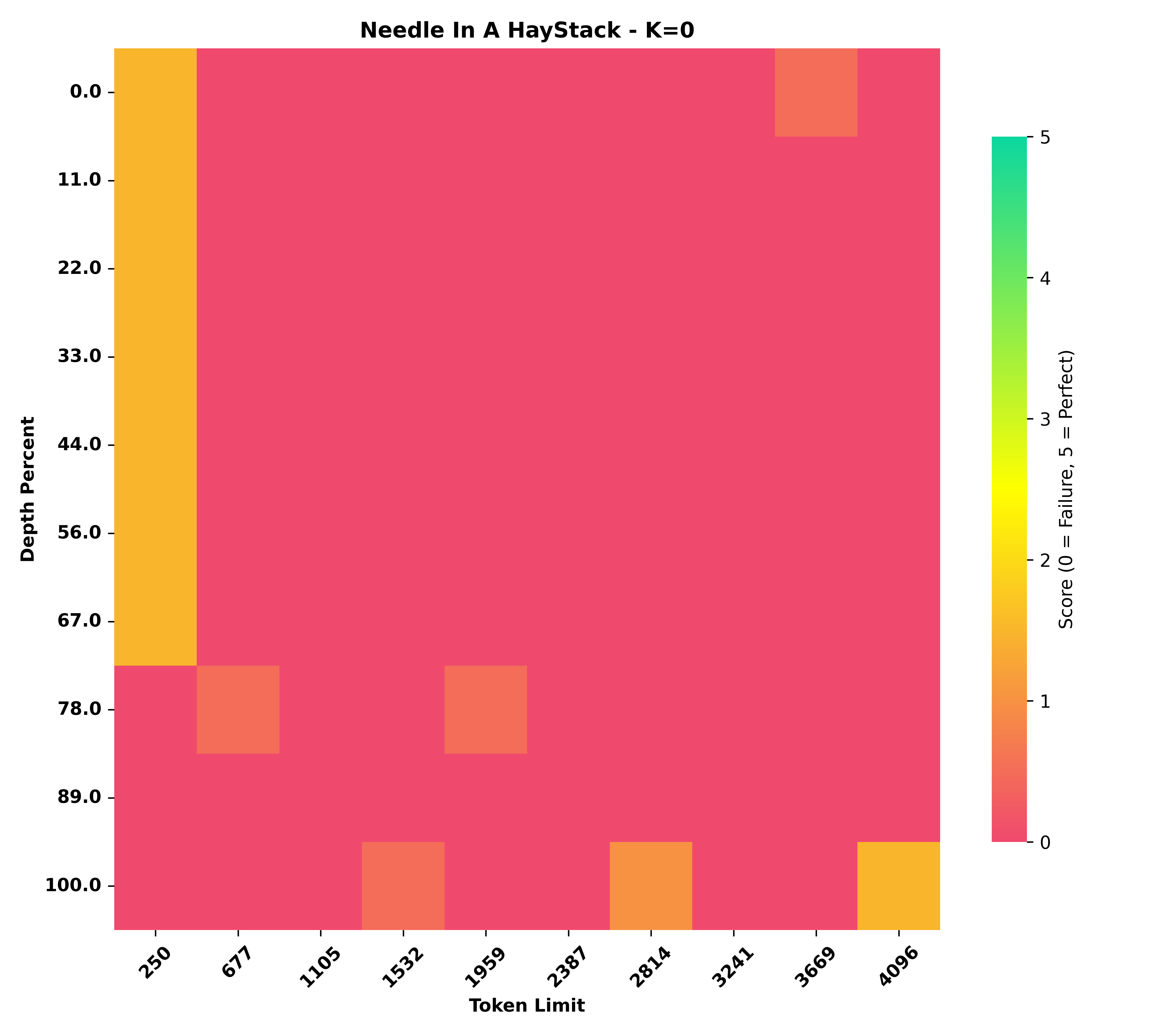}
    \caption{Jamba - Base}
    \label{fig:jamba_k0}
\end{subfigure}

\begin{subfigure}{0.3\linewidth}
    \centering
    \includegraphics[width=0.8\linewidth]{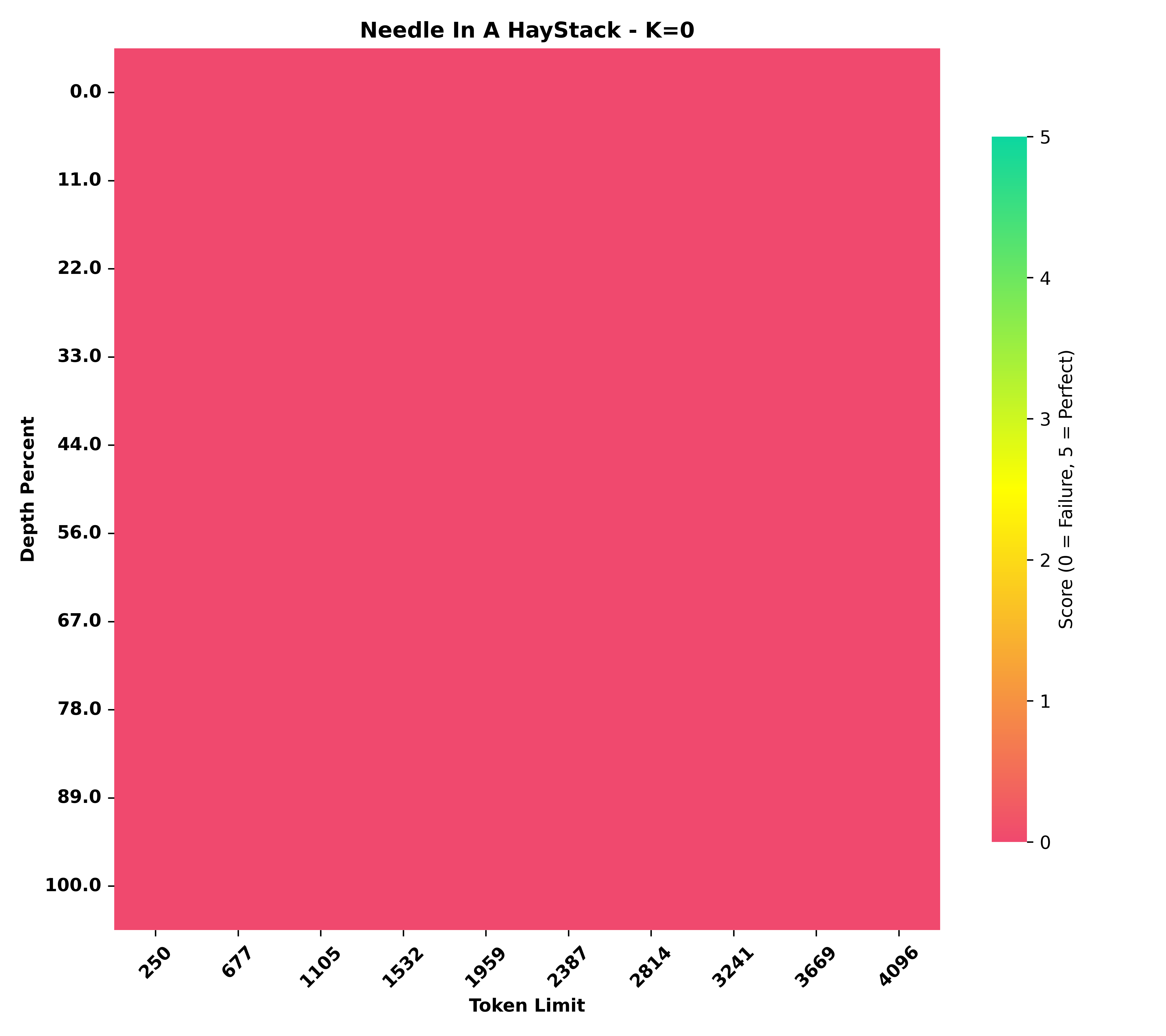}
    \caption{RG-2B - JRT}
    \label{fig:jrt_rg2b_k0}
\end{subfigure}
\begin{subfigure}{0.3\linewidth}
    \centering
    \includegraphics[width=0.8\linewidth]{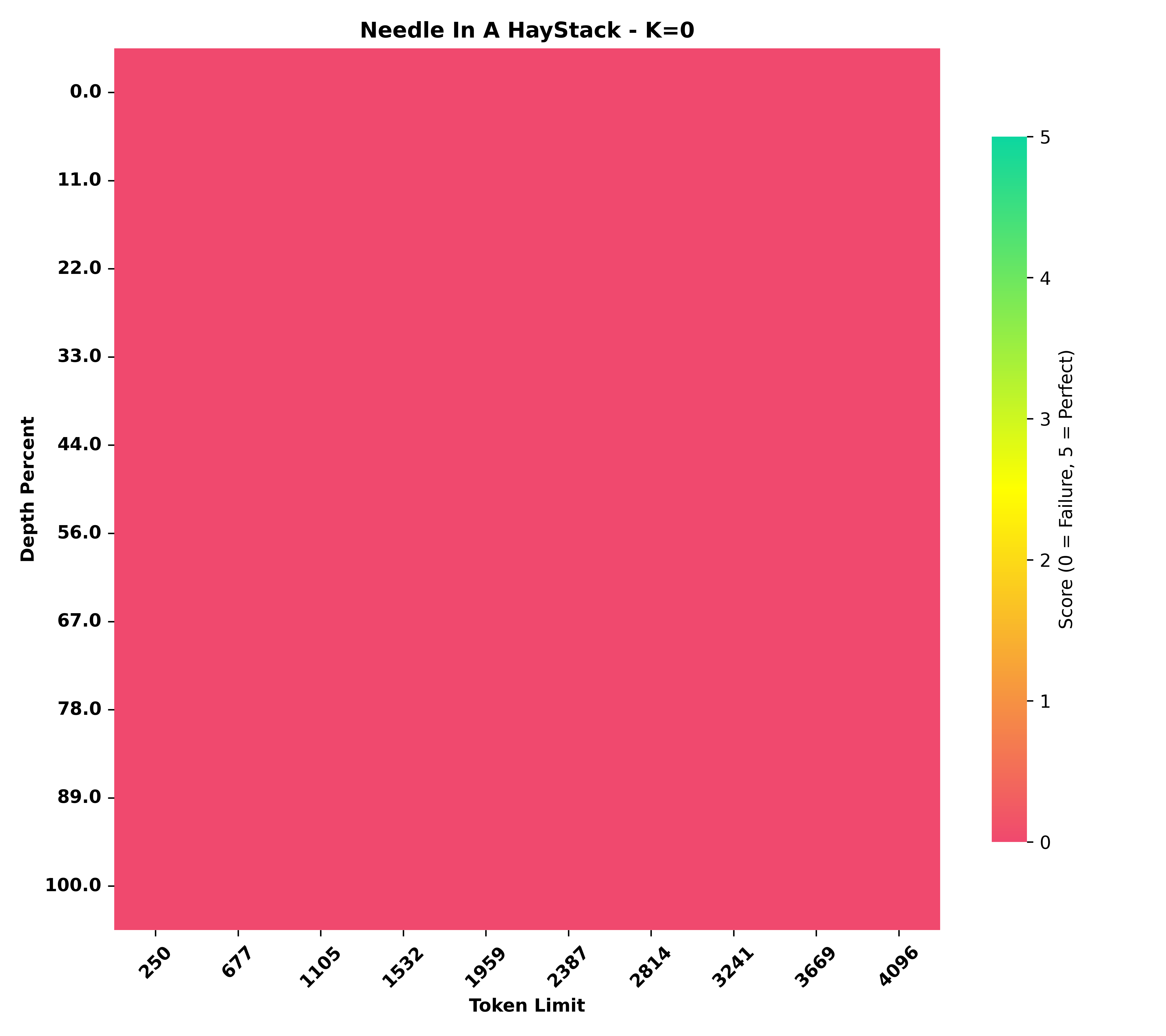}
    \caption{RG-9B - JRT}
    \label{fig:jrt_rg9b_k0}
\end{subfigure}
\begin{subfigure}{0.3\linewidth}
    \centering
    \includegraphics[width=0.8\linewidth]{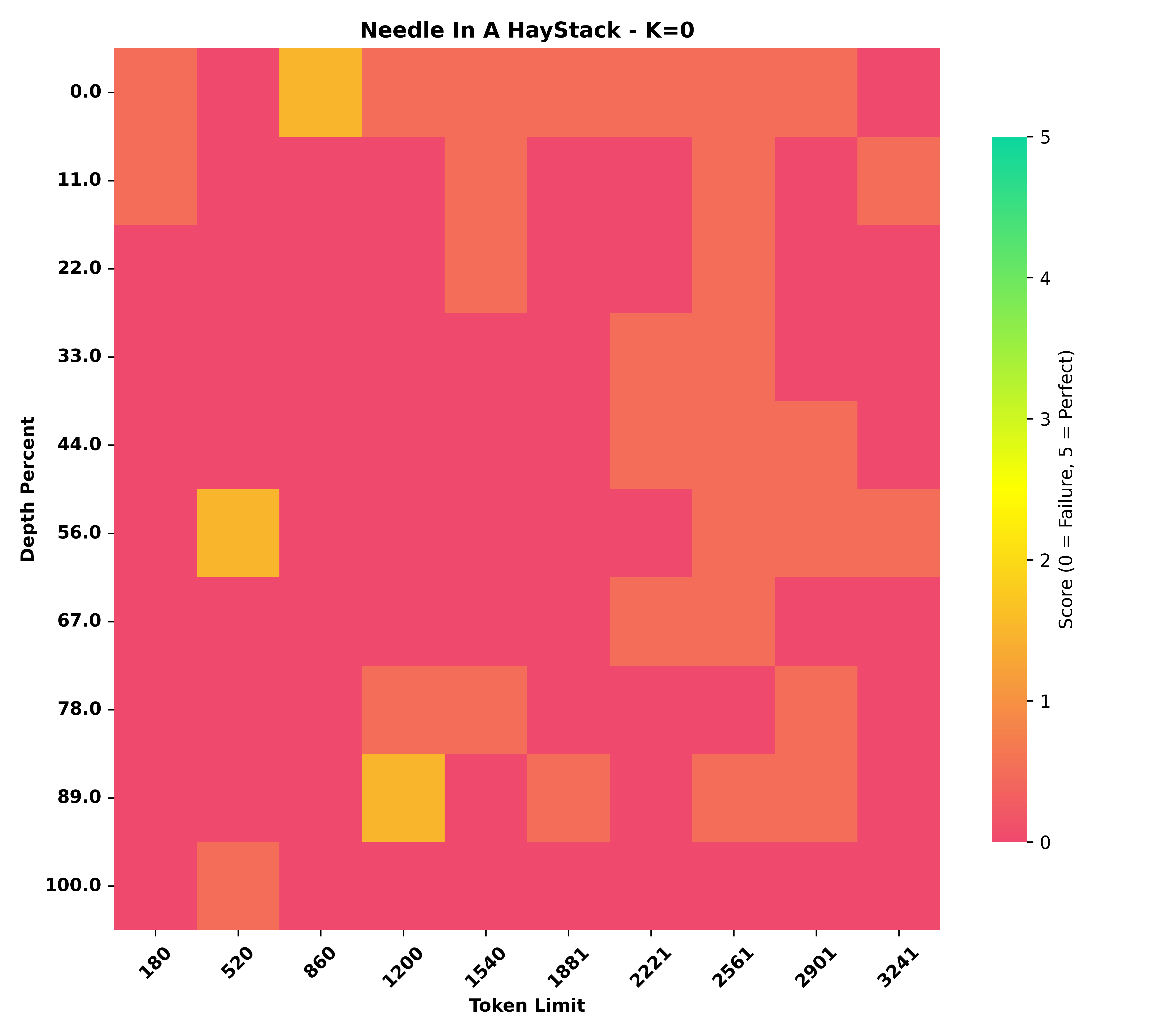}
    \caption{Jamba - JRT}
    \label{fig:jrt_jamba_k0}
\end{subfigure}
\caption{Retrieval maps for RG-2B, RG-9B and Jamba  at $k=0$ on NIAH with and without JRT applied.}
\label{fig:jrt_compare_k0}
\end{figure}

\begin{figure}[H]
\centering
\begin{subfigure}{0.3\linewidth}
    \centering
    \includegraphics[width=0.8\linewidth]{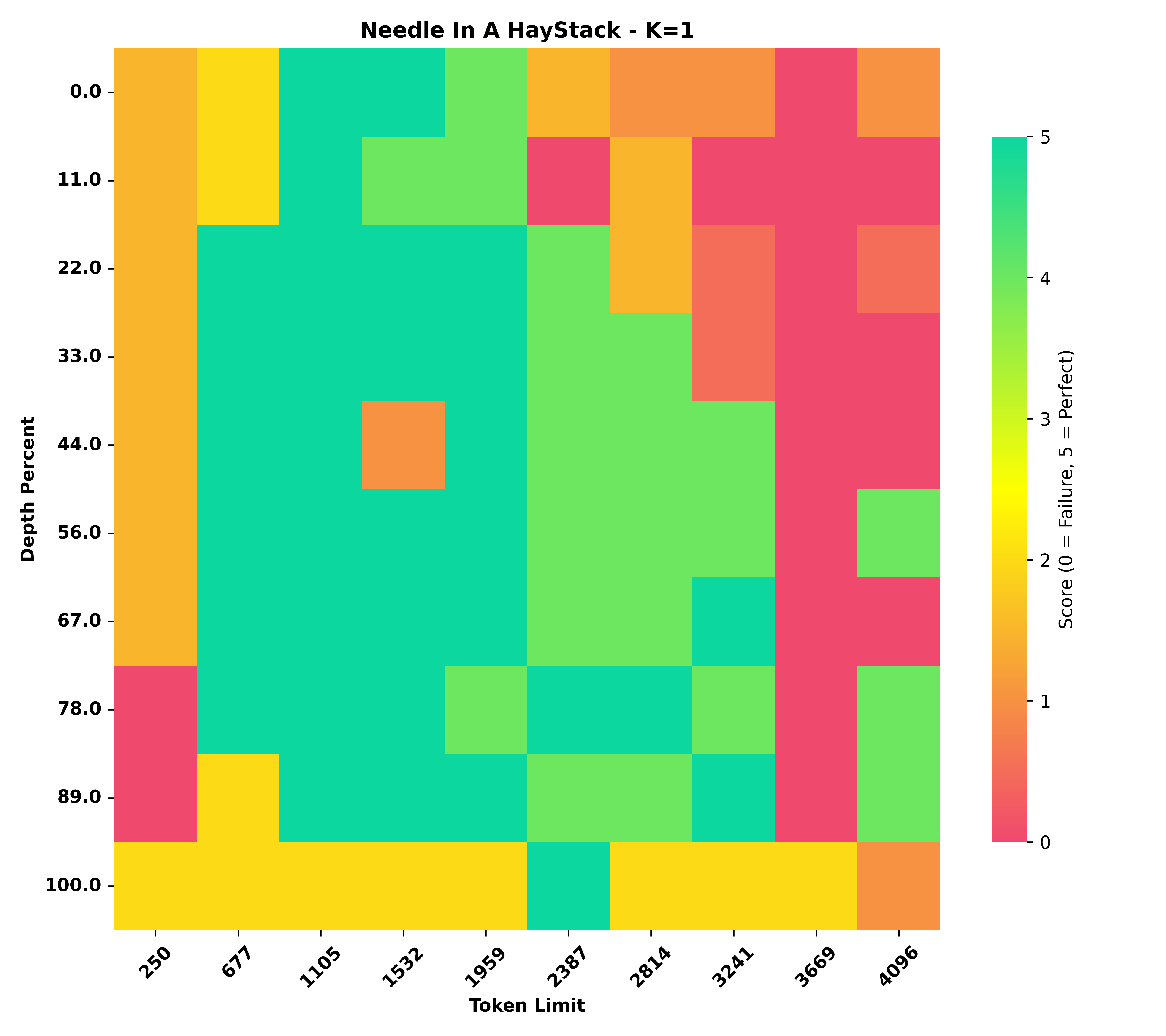}
    \caption{RG-2B - Base}
    \label{fig:rg2b_k1}
\end{subfigure}
\begin{subfigure}{0.3\linewidth}
    \centering
    \includegraphics[width=0.8\linewidth]{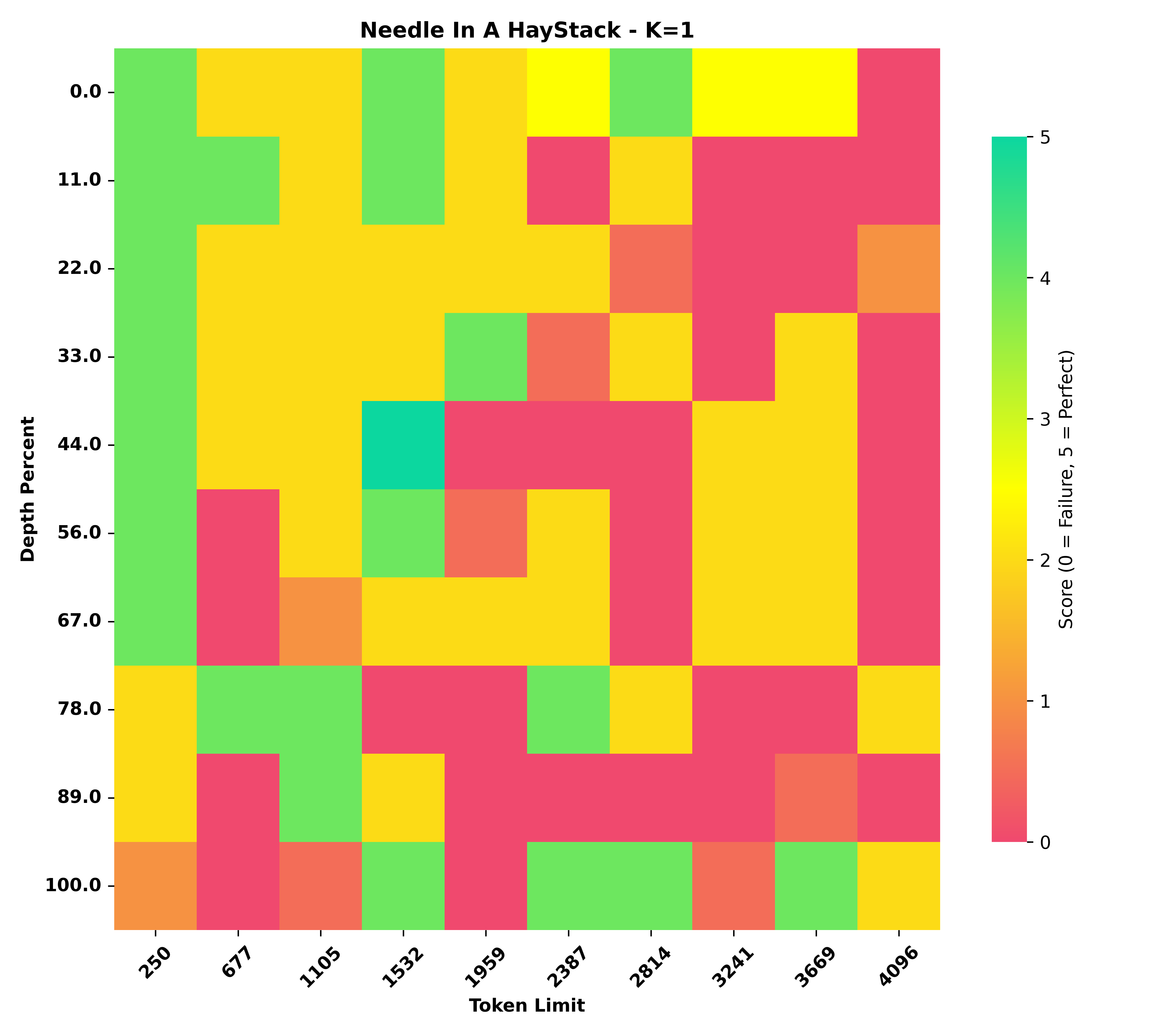}
    \caption{RG-9B - Base}
    \label{fig:rg9b_k1}
\end{subfigure}
\begin{subfigure}{0.3\linewidth}
    \centering
    \includegraphics[width=0.8\linewidth]{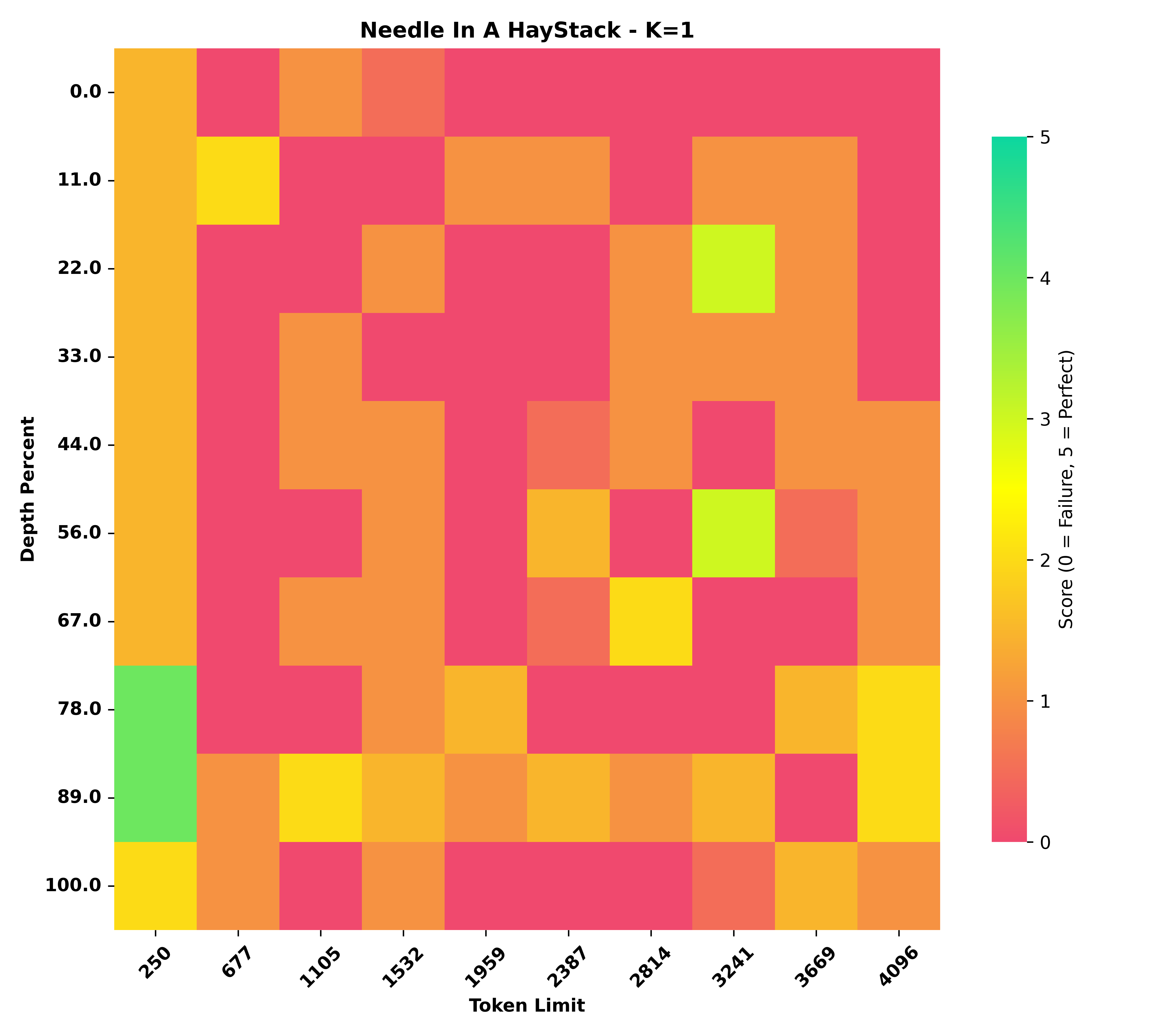}
    \caption{Jamba - Base}
    \label{fig:jamba_k1}
\end{subfigure}

\begin{subfigure}{0.3\linewidth}
    \centering
    \includegraphics[width=0.8\linewidth]{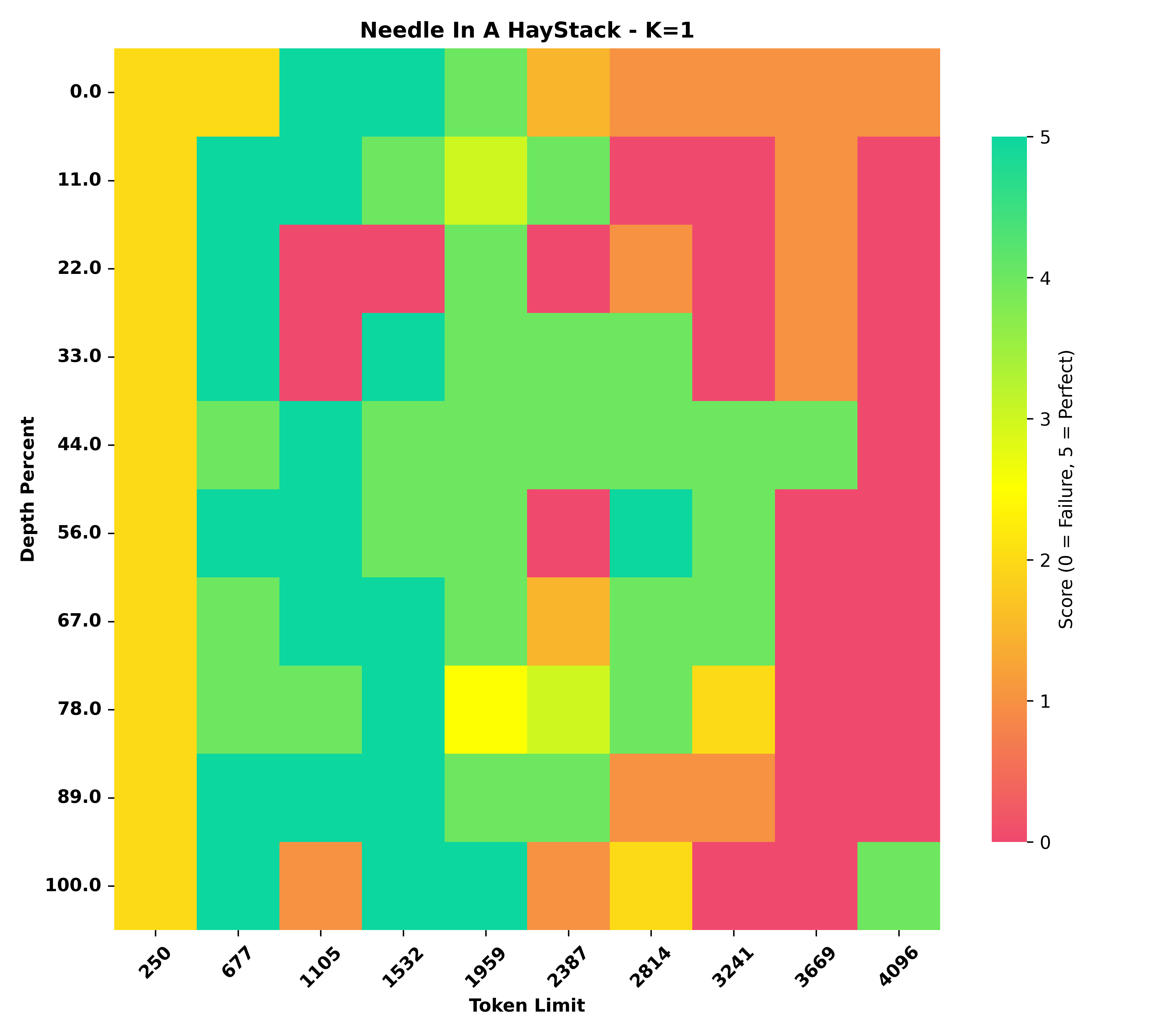}
    \caption{RG-2B - JRT}
    \label{fig:jrt_rg2b_k1}
\end{subfigure}
\begin{subfigure}{0.3\linewidth}
    \centering
    \includegraphics[width=0.80\linewidth]{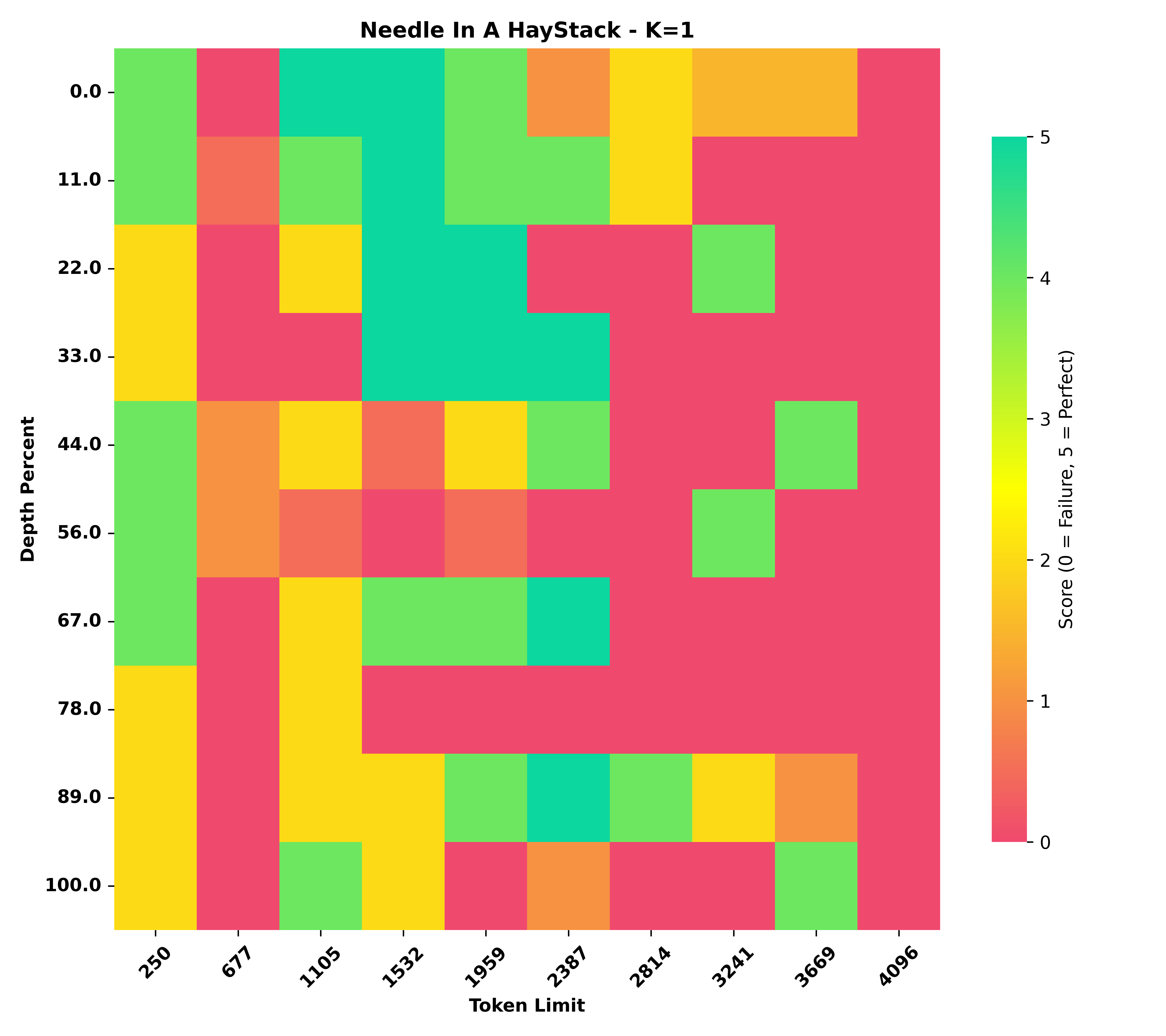}
    \caption{RG-9B - JRT}
    \label{fig:jrt_rg9b_k1}
\end{subfigure}
\begin{subfigure}{0.3\linewidth}
    \centering
    \includegraphics[width=0.80\linewidth]{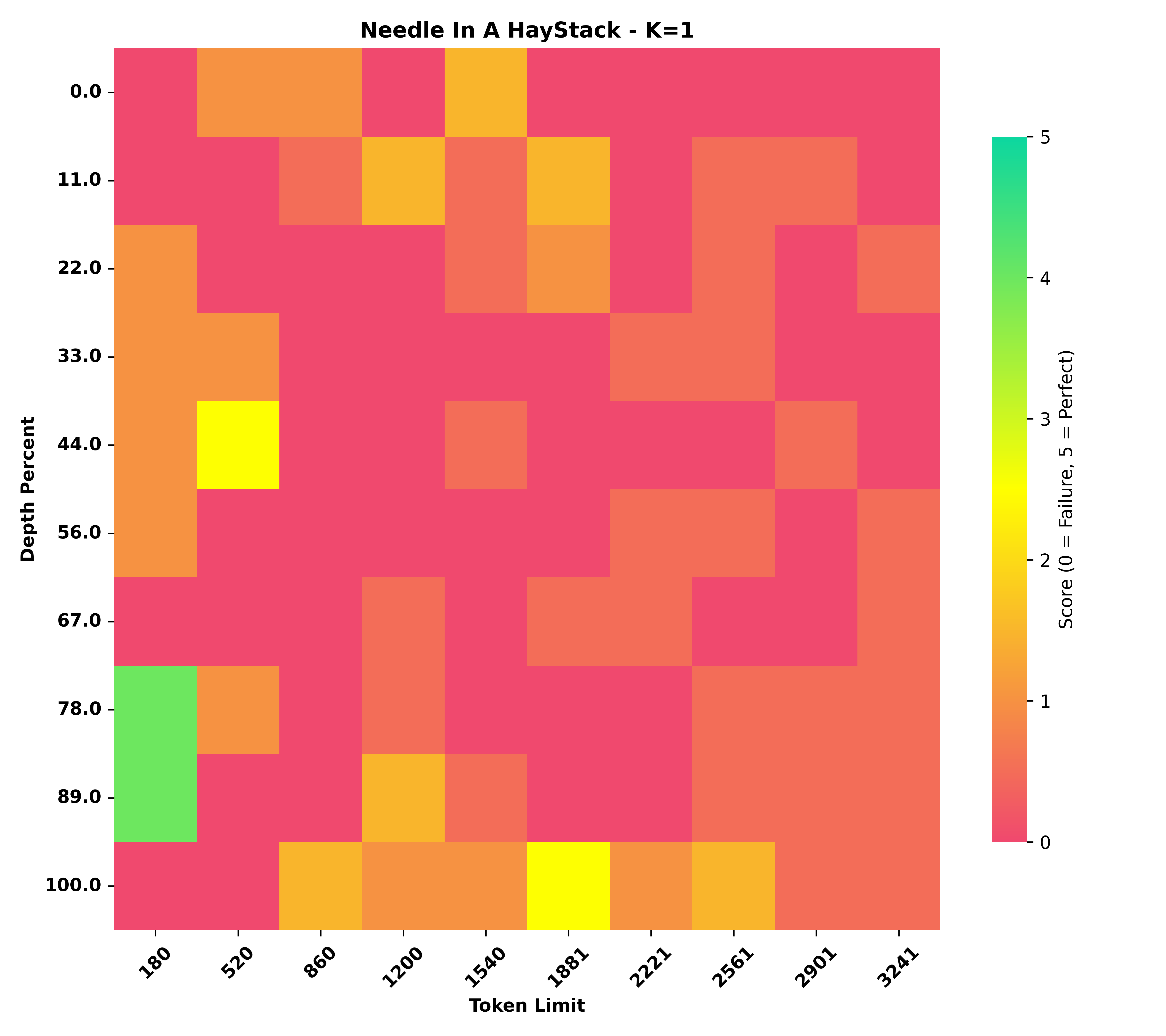}
    \caption{Jamba - JRT}
    \label{fig:jrt_jamba_k1}
\end{subfigure}
\caption{Retrieval maps for RG-2B, RG-9B and Jamba  at $k=1$ on NIAH with and without JRT applied.}
\label{fig:jrt_compare_k1}
\end{figure}

\begin{figure}[H]
\centering
\begin{subfigure}{0.3\linewidth}
    \centering
    \includegraphics[width=0.8\linewidth]{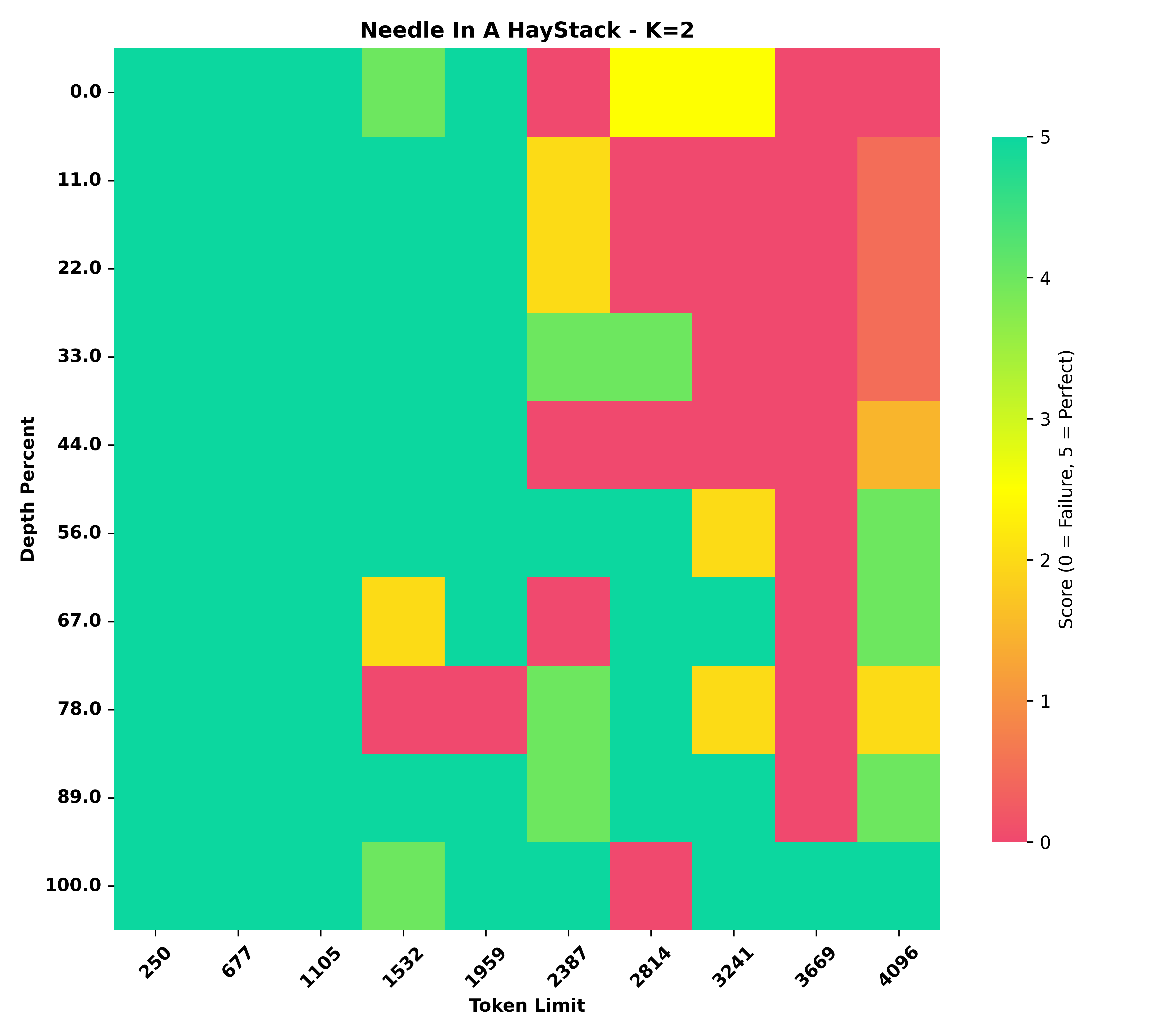}
    \caption{RG-2B - Base}
    \label{fig:rg2b_k2}
\end{subfigure}
\begin{subfigure}{0.3\linewidth}
    \centering
    \includegraphics[width=0.8\linewidth]{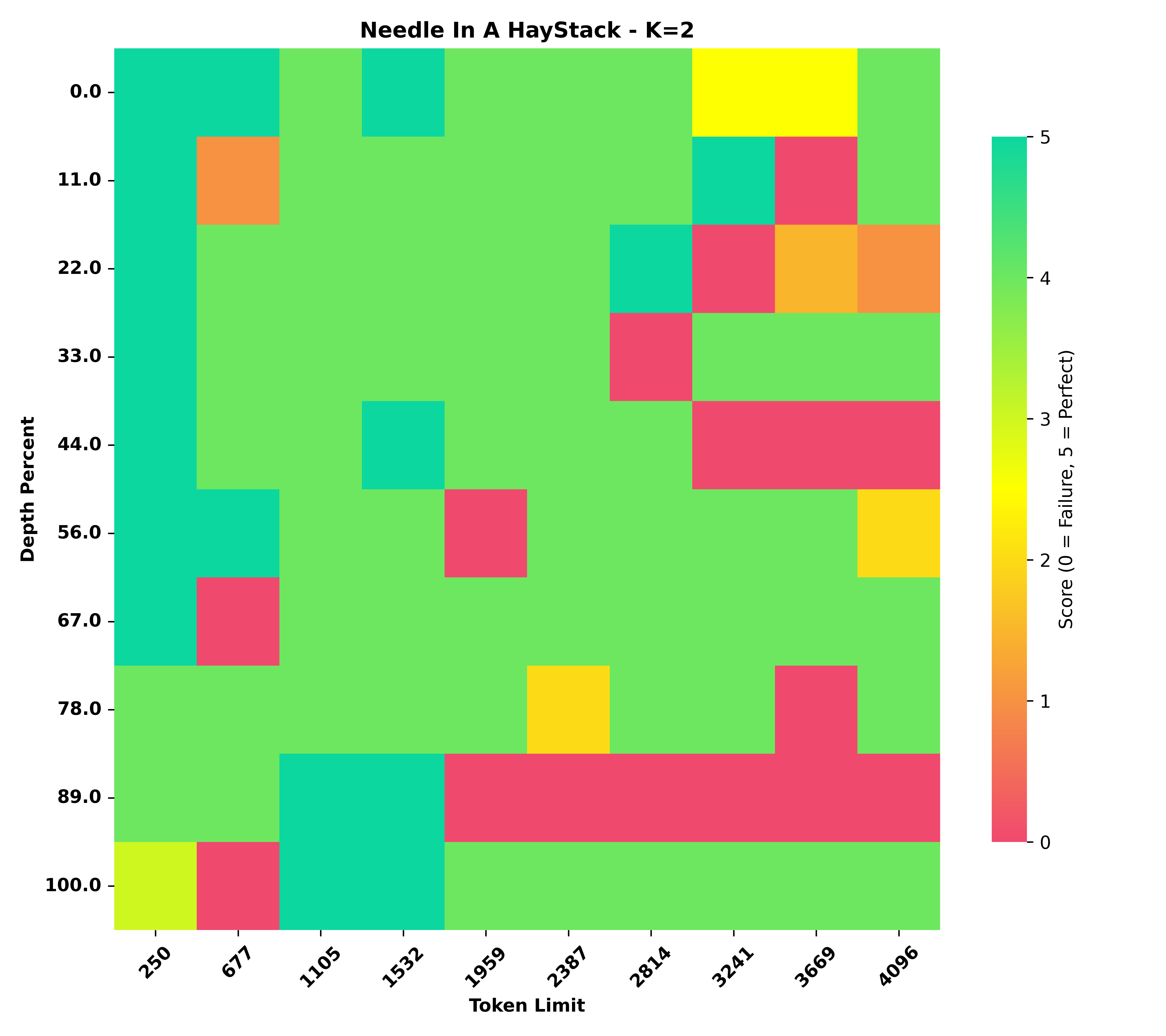}
    \caption{RG-9B - Base}
    \label{fig:rg9b_k2}
\end{subfigure}
\begin{subfigure}{0.3\linewidth}
    \centering
    \includegraphics[width=0.8\linewidth]{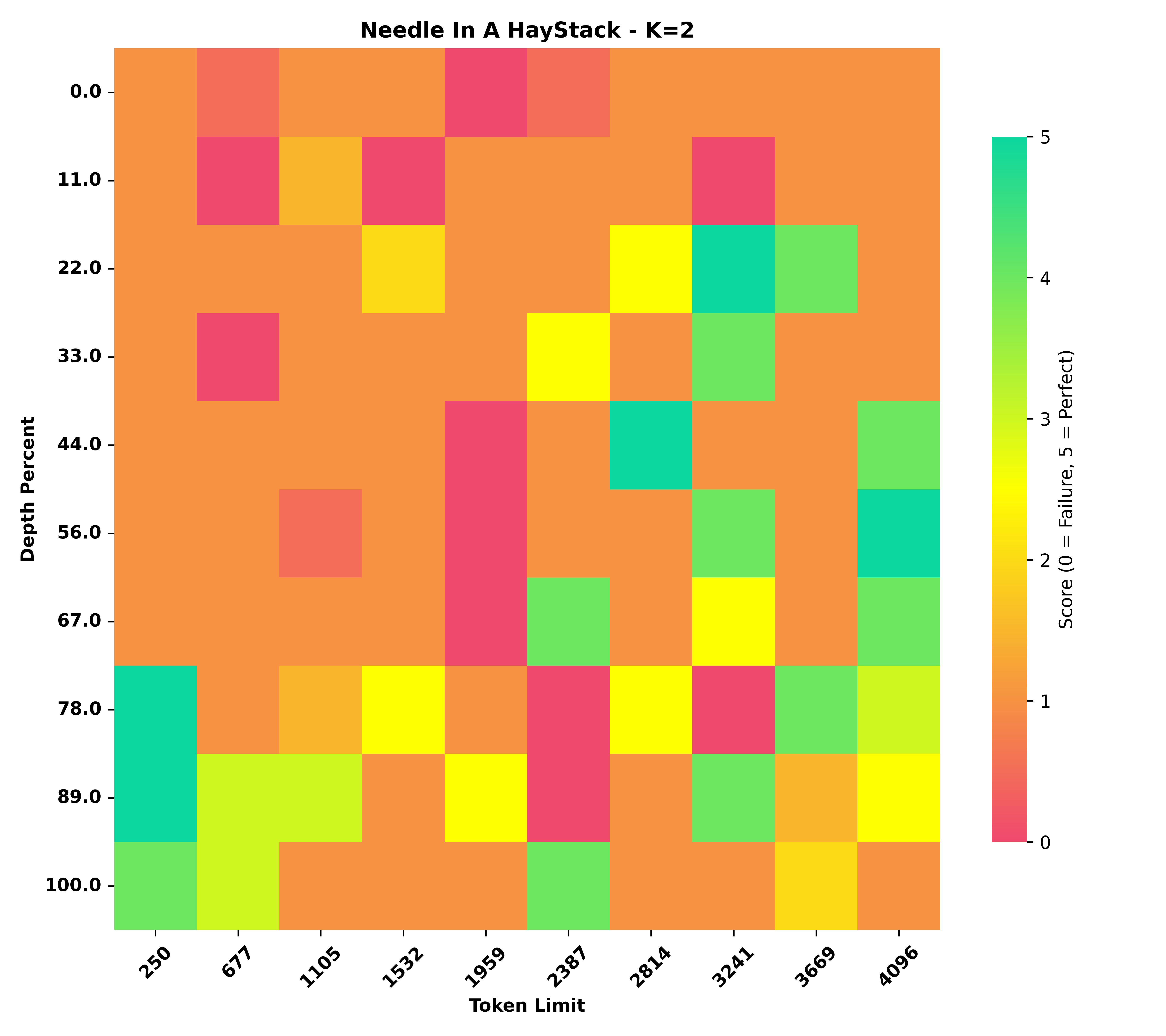}
    \caption{Jamba - Base}
    \label{fig:jamba_k2}
\end{subfigure}

\begin{subfigure}{0.3\linewidth}
    \centering
    \includegraphics[width=0.8\linewidth]{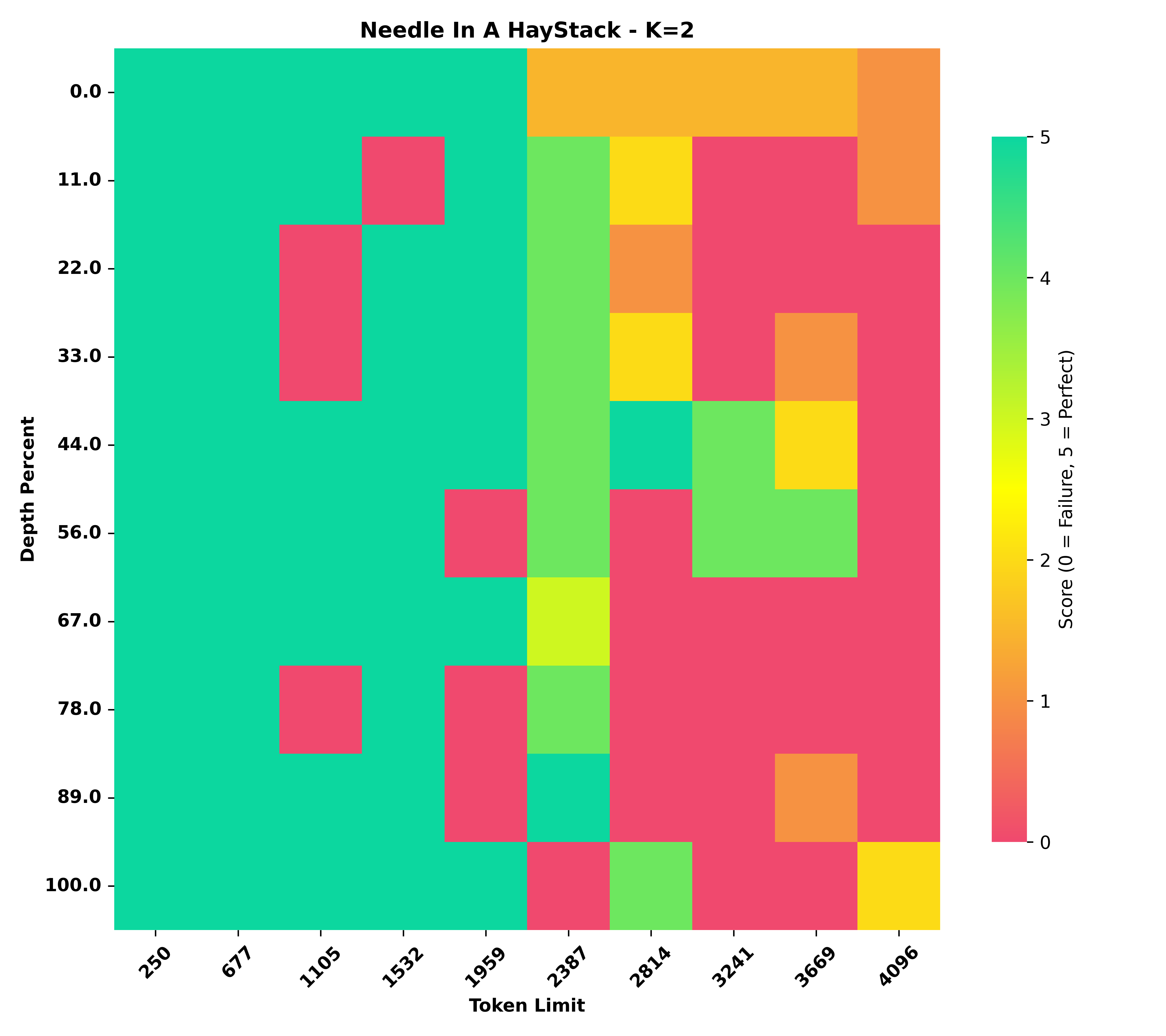}
    \caption{RG-2B - JRT}
    \label{fig:jrt_rg2b_k2}
\end{subfigure}
\begin{subfigure}{0.3\linewidth}
    \centering
    \includegraphics[width=0.8\linewidth]{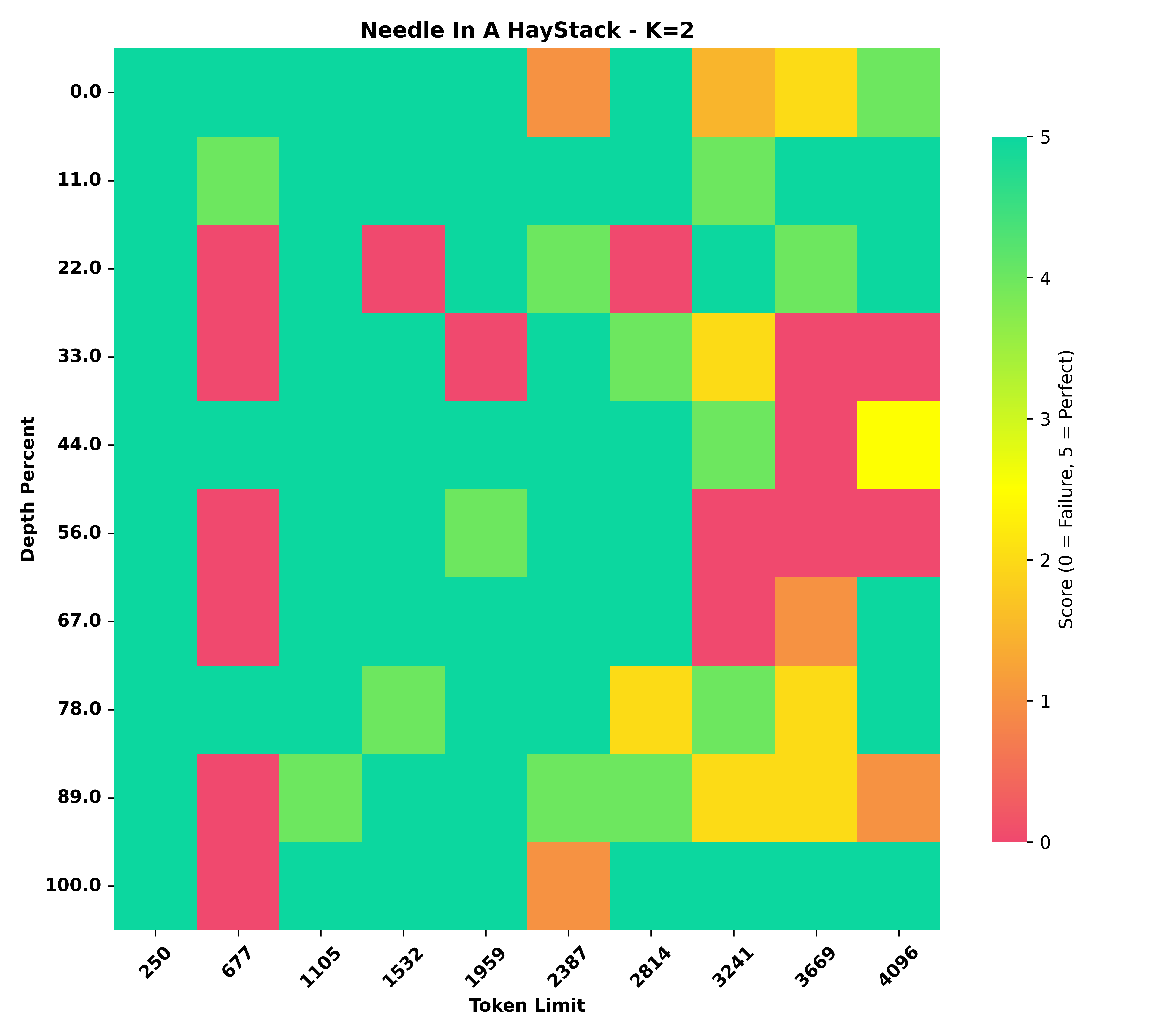}
    \caption{RG-9B - JRT}
    \label{fig:jrt_rg9b_k2}
\end{subfigure}
\begin{subfigure}{0.3\linewidth}
    \centering
    \includegraphics[width=0.8\linewidth]{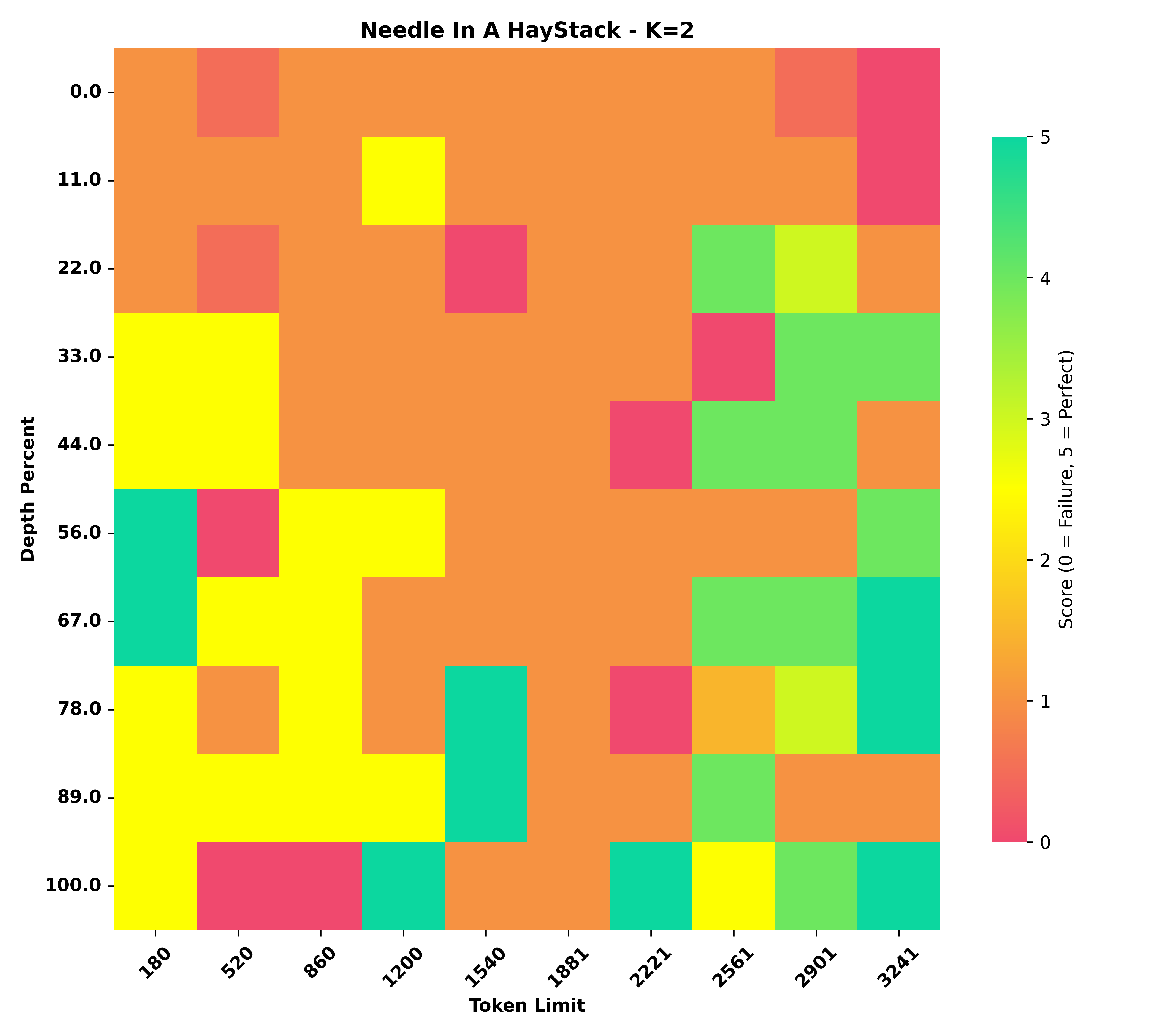}
    \caption{Jamba - JRT}
    \label{fig:jrt_jamba_k2}
\end{subfigure}
\caption{Retrieval maps for RG-2B, RG-9B and Jamba  at $k=2$ on NIAH with and without JRT applied.}
\label{fig:jrt_compare_k2}
\end{figure}

In Section \ref{q1}, we are also testing if applying JRT improves retrieval capabilities for top-k-sparsified RG-2B, RG-9B and Jamba. Figure \ref{fig:jrt_compare_k0}, \ref{fig:jrt_compare_k1}, and \ref{fig:jrt_compare_k2} compare the retrieval maps of standard NIAH and JRT-applied NIAH on the same $k$ on all models. Refer to Figure \ref{fig:manipulation} for an explanation of the axes.

\section{LME Benchmark}\label{lme}
\begin{figure}[h!]
    \centering
    \includegraphics[width=0.95\linewidth]{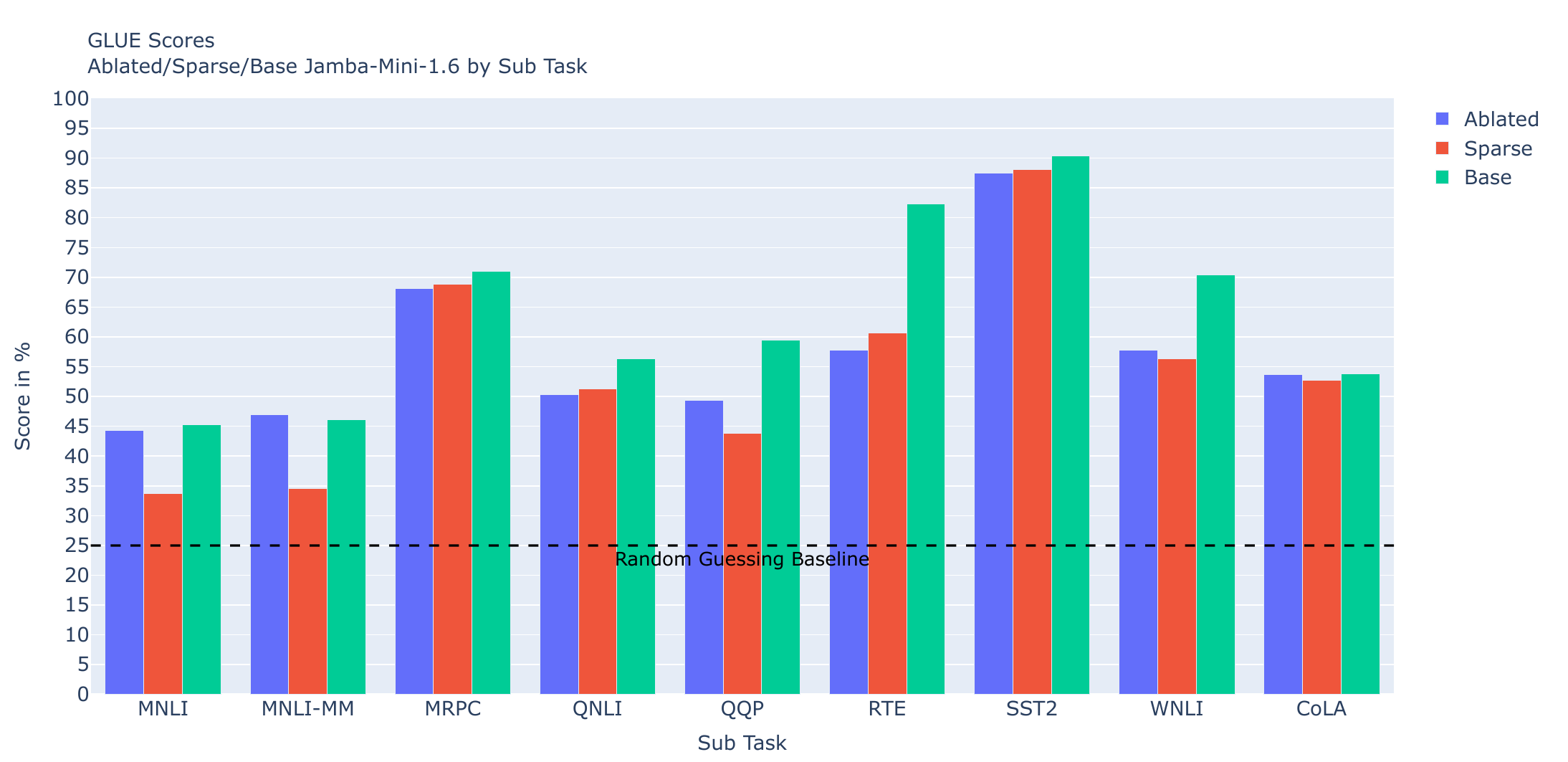}
    \caption{GLUE benchmark Sub Task scores of the ablated ($k=0)$, sparse ($k=5$) and base configuration.}
    \label{fig:detailed_glue}
\end{figure}

\begin{figure}[h!]
    \centering
    \includegraphics[width=0.95\linewidth]{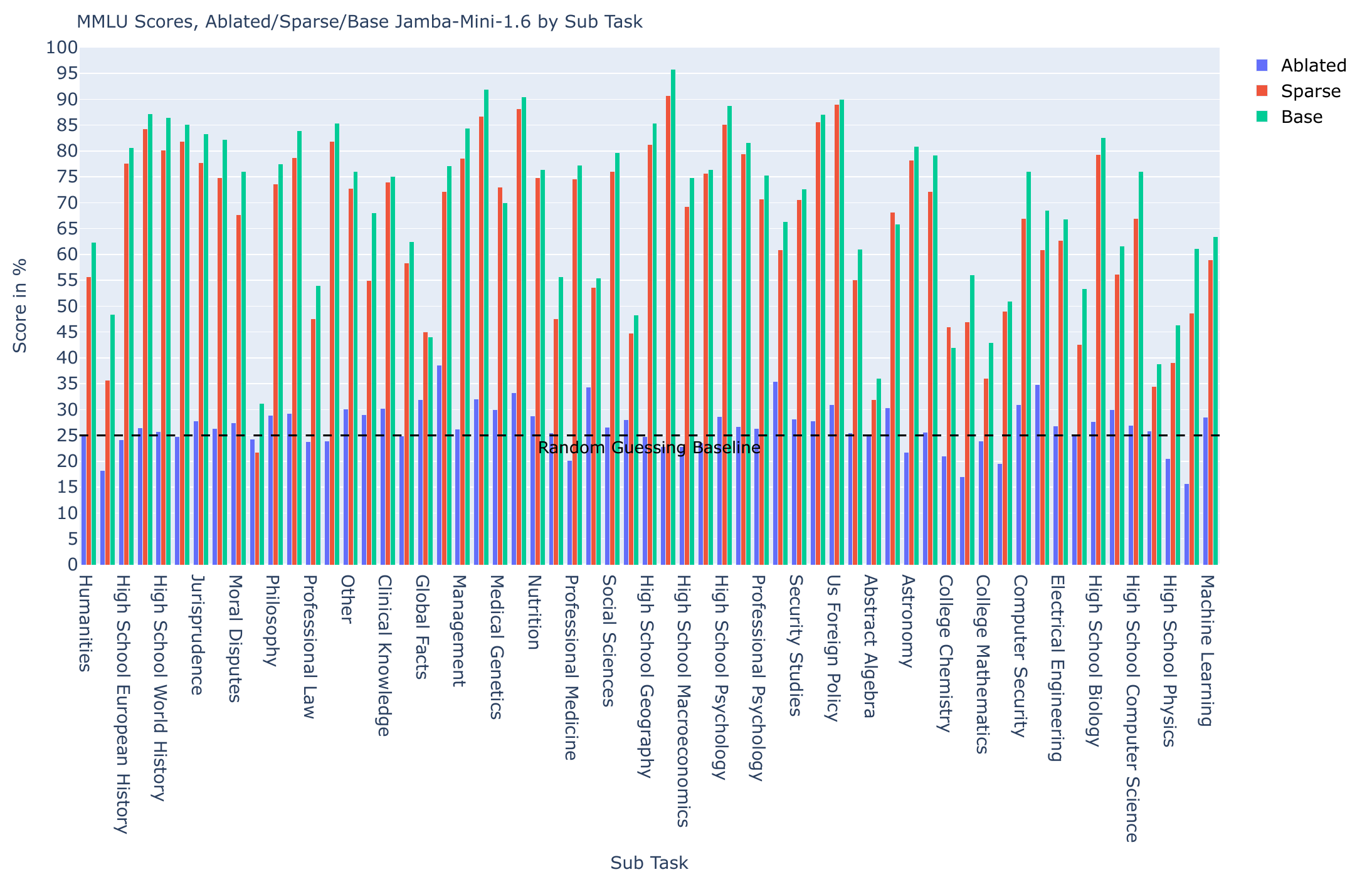}
    \caption{MMLU benchmark Sub Task scores of the ablated ($k=0)$, sparse ($k=5$) and base configuration.}
    \label{fig:detailed_mmlu}
\end{figure}

In Section \ref{q3}, we tested Jamba using LME. It is an instruction-tuned model, so evaluation was run with the \texttt{apply\_chat\_template} and \texttt{fewshot\_as\_multiturn} flags. Figures \ref{fig:detailed_glue} and \ref{fig:detailed_mmlu} show the performance of Jamba in the ablated, sparsified and base version per subtask on GLUE and MMLU.

\section{Retrieval maps for attention weight manipulation}\label{attention-manipulation}
\begin{figure}[H]
    \centering
    \includegraphics[width=0.8\linewidth]{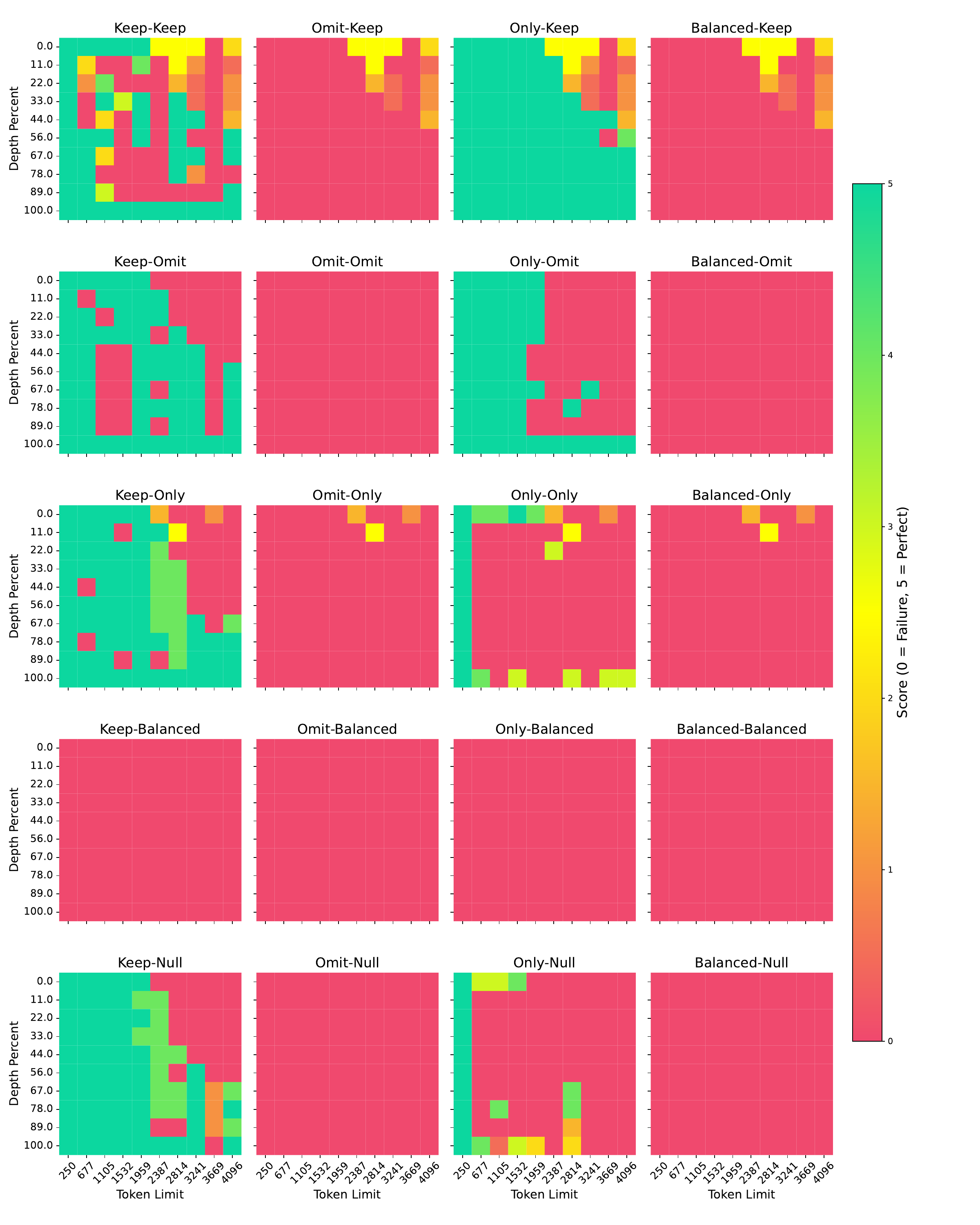}
    \caption{Retrieval maps for all manipulation combinations used in Section \ref{q4}.}
    \label{fig:manip_total}
\end{figure}

In Section \ref{q4}, we systematically manipulated the attention weights in RG-2B during NIAH benchmarks. Figure \ref{fig:manip_total} shows the retrieval maps of those benchmarks, structured in the same way as Table \ref{tab:man_res} for easy comparison.

\section{Pearls of Wisdom}\label{memos}

Investigating and experimenting with manipulated models often results in amusing outputs. Since life is about joy, and reading some of those outputs definitely caused us joy, we distilled a collection of output strings to share our laughter with the world. Some quotes are also absurdly wise, which led us to believe they were part of the Paul Graham essays that were used to fill the context in the NIAH benchmark. However, none of the generated strings were found in the context strings. All following quotes were generated by sparsified or manipulated RG-2B or RG-9B, in response to the question: "What is the best thing to do in San Francisco?"

\setlength{\epigraphwidth}{1.0\linewidth}

\epigraph{
The best way to get to the top of the mountain is to start by walking.

- Mark Twain

The first thing I want to say is that I am not a native English speaker
}{RG-2B - Binary-Keep}

\epigraph{"The most important thing is to be kind. It is the little things that make the difference."

- Unknown}{RG-2B - Binary-Keep}

\epigraph{The first sentence is a sentence that is not a sentence. It is a sentence that
has been written in a way that is not a sentence.}{RG-2B - Binary-Null}

\epigraph{"The best thing to do in San Francisco is eat a sandwich and sit in Dolores Park on a sunny day on a sunny day."

The above is a joke.}{RG-2B - Only-Keep} 

\epigraph{I am a programmer. I am a programmer. I am a programmer. I am a programmer. I am a programmer. I am a programmer. I am a programmer.}{RG-2B - $k=0$, sparse prefill}

\epigraph{You need humility to know when to use qualification.}{RG-9B - $k=4$} 

\end{document}